# Hybrid of representation learning and reinforcement learning for dynamic and complex robotic motion planning

Chengmin Zhou, Xin Lu, Jiapeng Dai, Bingding Huang, Xiaoxu Liu, and Pasi Fränti

*Abstract*—Motion planning is the soul of robot decision making. Classical planning algorithms like graph search and reaction-based algorithms face challenges in cases of dense and dynamic obstacles. Deep learning algorithms generate suboptimal one-step predictions that cause many collisions. Reinforcement learning algorithms generate optimal or near-optimal time-sequential predictions. However, they suffer from slow convergence, suboptimal converged results, and overfittings. This paper introduces a hybrid algorithm for robotic motion planning: long short-term memory (LSTM) pooling and skip connection for attention-based discrete soft actor critic (LSA-DSAC). First, graph network (relational graph) and attention network (attention weight) interpret the environmental state for the learning of the discrete soft actor critic algorithm. The expressive power of attention network outperforms that of graph in our task by difference analysis of these two representation methods. However, attention based DSAC faces the overfitting problem in training. Second, the skip connection method is integrated to attention based DSAC to mitigate overfitting and improve convergence speed. Third, LSTM pooling is taken to replace the sum operator of attention weigh and eliminate overfitting by slightly sacrificing convergence speed at early-stage training. Experiments show that LSA-DSAC outperforms the state-of-the-art in training and most evaluations. The physical robot is also implemented and tested in the real world.

*Index Terms*—Motion Planning, Navigation, Reinforcement Learning, Representation Learning, Intelligent Robot

## I. INTRODUCTION

Intelligent robots play an important role in our daily life. For example, autonomous robot has been applied to hotel guidance [1], parcel delivery [2][3], and robotic arms in manufacturing [4][5]. Motion planning or path planning is the soul of robotic decision making. It enables robots to reach the goal and finish the tasks.

Classical planning algorithms like graph search (e.g., A* [6]) enable robots to navigate in static environment. However, they cause many collisions in the environment with dense and dynamic obstacles because of the huge burden in updating the environmental map in real time.

Classical reaction-based algorithms like *dynamic window approach* (DWA) [7] and *optimal reciprocal collision avoidance* (ORCA) [8] reduce collisions in the environment with dense and dynamic obstacles because they compute robot's motion by just considering obstacle's geometry and speed information. This requires fewer information updates compared to the map update. However, high collision rates still exist in reaction-based algorithms when the robot avoids obstacles with high speed because of the increasing burden in the information update.

*Deep learning algorithms* (DL) like *convolutional neural network* (CNN) [9] avoid the information update problem of the reaction-based algorithms by training a model that generates robot's decisions or actions in real time. However, decisions from DL are based on one-step predictions which result in suboptimal trajectories when the robot moves toward its goals.

*Reinforcement learning* (RL) algorithms like *deep Q network* (DQN) [10] and *advantage actor critic* (A2C) or *asynchronous advantage actor critic* (A3C) [11] improve the one-step predictions of CNN by training the models based on the multi-step time-sequential predictions. These multi-step time-sequential predictions are better than the one-step predictions because the training of the RL model considers the time-sequential information of the goals and obstacles. RL, however, may suffer from slow convergence speed and suboptimal converged result once the input (the environment state) is low-quality with limited expressive power.

**Progress of representation learning and RL.** Currently, representation learning methods like LSTM pooling [12], graph network [13][14], and attention network [15][16] improve the expressive power of input. RL algorithms are also improved by introducing new architectures like double Q networks [17], dueling architecture [18], and deterministic architecture [19][20]. The fused architecture of the double Q networks and the actor-critic network is proved to be one of the most efficient architectures in RL [21]. This architecture is also applied to many RL variants like *Twin delayed deep deterministic policy gradient* (TD3) [22] and *soft actor critic* (SAC) [23][24][25]. The combination of representation learning and RL is a promising

First correspondence: Pasi Fränti (email: franti@cs.uef.fi). Affiliation: Machine Learning Group, School of Computing, University of Eastern Finland, Joensuu, Finland.
Second correspondence: Bingding Huang (email: huangbingding@sztu.edu.cn). Affiliation: College of Big Data and Internet, Shenzhen Technology University, Shenzhen, China.
First author: Chengmin Zhou (email: zhou@cs.uef.fi). Chengmin Zhou is with University of Eastern Finland and Shenzhen Technology University simultaneously.
Second author: Xin Lu (email: luwenkai67109@outlook.com). Affiliation: Sino-German College of Intelligent Manufacturing, Shenzhen Technology University, Shenzhen, China.
Third author: Jiapeng Dai (email: jalaxy1996@outlook.com). Affiliation: College of Big Data and Internet, Shenzhen Technology University, Shenzhen, China.
Fifth author: Xiaoxu Liu (email: liuxiaoxu@sztu.edu.cn). Affiliation: Sino-German College of Intelligent Manufacturing, Shenzhen Technology University, Shenzhen, China.

direction for better motion planning performance, because RL is fed with the input with high expressive power. This improves the overall convergence of RL algorithms.

**Technical difficulties of existing works.** The combination of representation learning and RL is promising for improving robotic motion planning performance. However, current works in this direction are still not good enough for challenging commercial tasks. Existing works about the combination of the representation learning and RL include the *relational graph* (RG) [13], *proximal policy optimization* (PPO) with multiple robots [26], CADRL [27], LSTM-A2C [28][29], LSTM-RL [15] and SARL [15].

RG is the combination of relational graph and DQN. Relational graph describes the relationship of all agents (the robot and obstacles), instead of focusing on the robot-obstacle relationship. Relational graph partly and indirectly represents the robot-obstacle relationship, therefore its expressive power is limited. DQN faces over-estimation problems which cause the slow and suboptimal convergence of overall networks.

PPO with multiple robots faces problems of data quality because it learns obstacle features from entire source environmental state without precisely and explicitly analyzing the relationship between the robot and obstacles. Moreover, the entire source environmental state is interpreted by the CNN in the PPO with multiple robots. Background noise is also included in this interpretation process, resulting in poor quality of interpreted environmental state.

CADRL learns the pairwise feature of the robot and one obstacle by DQN. Then, a trained model is applied to the multiple-obstacle case. CADRL is myopic because it does not consider the relationship between the robot and obstacle. The closest obstacle feature is just used for training instead of all obstacle features. DQN in CADRL also brings high bias and variance.

In LSTM-A2C and LSTM-RL, LSTM encodes obstacle features by distance-based order which partly represents robot-obstacle relationship, resulting in limited expressive power of interpreted environmental state. A2C/A3C lack efficient data replay strategies, resulting in a slow convergence. A2C/A3C and DQN in LSTM-A2C and LSTM-RL bring high bias and variance, resulting in a slow convergence.

SARL consists of an attention network and DQN where attention network interprets robot-obstacle features to the attention weight that better describes the relationship between the robot and obstacles, resulting in the improvement of expressive power. However, attention network still faces the overfitting problem if overall architecture has deep and complex networks. Moreover, DQN brings high bias and variance. These two reasons cause the slow and suboptimal convergence of SARL.

**Optimizations and contributions.** For better motion planning performance of the robot among dense and dynamic obstacles,

1) we first implemented the *discrete action for soft actor critic* (DSAC) which is the soft actor critic algorithm in the setting of discrete action space, and is also one of most efficient RL algorithms currently. DSAC is then combined with the *relational graph* (RG) [13], resulting in the *relational graph based DSAC* (RG-DSAC) that achieves satisfactory performance in motion planning. However, we found that the expressive power of the relational graph is limited in the experiment. The relational graph just partly describes the relationship between the robot and obstacles via establishing the relationships for all agents without precisely focusing on the relationship between the robot and obstacles. This may result in the limited expressive power of interpreted environmental state.

2) The expressive power of interpreted environmental state is improved by replacing the relational graph using *attention weight* (AW) [15] which precisely and explicitly analyze and describe the relationship between the robot and obstacles. This results in the *attention weight based DSAC* (AW-DSAC) which outperformed RG-DSAC in the early-stage training but suffered from overfittings.

3) After analysis, we concluded that the *feature loss* and *pooling method* in attention network may cause the overfitting. Hence, we optimized attention network by integrating the skip connection method and LSTM pooling into the architecture of the attention network, resulting in the *skip connection for attention-based DSAC* (SA-DSAC) and LSA-DSAC. SA-DSAC *mitigated* the problem of overfittings in training in case of fewer dynamic obstacles. LSA-DSAC *eliminated* overfittings by sacrificing the convergence speed slightly at the early-state training.

Overall, the workflow of our motion planning task is shown in Fig. 1. Main contributions of this paper include

1) the implementation of RG-DSAC and AW-DSAC,

2) the LSA-DSAC which is the optimized version of AW-DSAC by integrating the skip connection method and LSTM pooling into the architecture of the attention network of AW-DSAC,

3) extensive evaluations of our algorithms against the state-of-the-art, and

4) physical implementation and testing of the robot in real world.

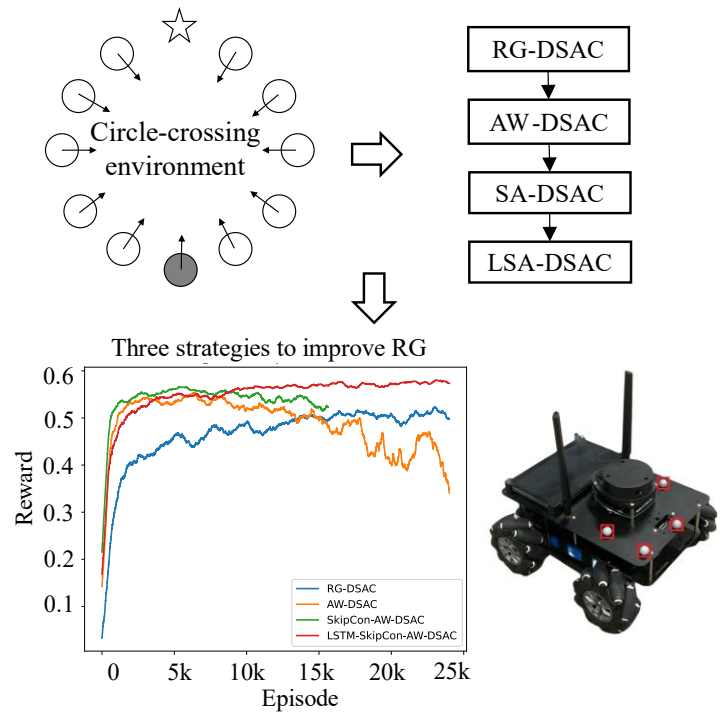

**Fig. 1.** Workflow of our LSA-DSAC. The training data is collected from the circle-crossing simulator. The relational graph based DSAC (RG-DSAC) is implemented and selected as the trainable baseline algorithm for comparisons. Relational graph is replaced the attention weight (or attention network), resulting in attention weight based DSAC (AW-DSAC). The skip connection is applied to the attention network to improve the convergence, resulting in the skip connection for Attention weight based DSAC

(SA-DSAC). Finally, LSTM is applied to the SA-DSAC to further improve the convergence, resulting in the LSTM encoding and skip connection for attention weight based DSAC (LSA-DSAC). The training curves demonstrate a good convergence of LSA-DSAC over the rest algorithms. This paper includes the physical implementation which demonstrates how to transplant our algorithm into the real world. Our test code is available on website https://github.com/CHUENGMINCHOU/LSA-DSAC.

This paper is arranged as follows: Section II presents the state-of-the-art, problem formulation and preliminary of RL and DSAC. Section III presents RG-DSAC, AW-DSAC, SA-DSAC and LSA-DSAC. Section IV presents network framework of LSA-DSAC, model trainings, model evaluations and physical implementation.

## II. RESEARCH BACKGROUND

This section first presents the state-of-the-art for dynamic robotic motion planning tasks. The state-of-the-art includes classical algorithm ORCA and trainable algorithms CADRL, LSTMRL, LSTM-A2C/A3C, PPO with multiple robots, SARL, and RG-DQN. Then, the problem formulation of motion planning tasks are given by mathematic descriptions. Finally, the preliminary of RL and DSAC are presented. They are fundamental concepts for following further algorithm implementations and optimizations.

### A. State-of-the-art for dynamic motion planning

This part concludes the state-of-the-art of motion planning algorithms. They includes ORCA [8], CADRL [27], LSTMRL [15][10], LSTM-A2C/A3C [11][29][28], PPO with multiple robots [30][26], SARL [15], RG-DQN [13]. Reaction-based ORCA relies on the positions and velocities of robots and obstacles to compute possible robot's velocity. CADRL is based on DQN that learns pairwise features of the robot and one obstacle. Trained model is then applied to multiple-obstacle cases. LSTMRL and LSTM-A2C/A3C are based on DQN and A2C/A3C to learn obstacle features that are pooled to hidden features by LSTM. PPO with multiple robots is based on CNN and PPO to learn from entire source environmental state that include features of the robot and obstacles and potential background noise. SARL is based on DQN where the attention network pools the pairwise robot-obstacle features to attention features (attention weight). RG-DQN is also based on DQN where the relation matrix and message passing process interpret source features to the graph features. We implemented ORCA, CADRL, LSTMRL, LSTM-A2C, SARL as baseline algorithms for comparisons.

### B. Problem formulation

All algorithms in this paper are trained and tested in simulators (Fig. 2) provided by ORCA [8]. Simulators includes *circle-crossing* and *square-crossing* simulators that add *predictable complexity* to the motion planning tasks. Let $a$ and $v$ represent the action and velocity of robot where $a = v = [v_x, v_y]$. Let $p = [p_x, p_y]$ represent the robot position. Let $s_t$ represent the robot state at time step $t$. $s_t$ consists of observable and hidden parts $s_t = [s_t^{obs}, s_t^h]$, $s_t \in R^9$. Observable part refers to factors that can be measured or observed by others. It consists of the position, velocity, and radius $s^{obs} = [p_x, p_y, v_x, v_y, r]$, $s^{obs} \in R^5$. The hidden part refers to factors that cannot be seen by others. It consists of planned goal position, preferred speed and heading angle $s^h = [p_{gx}, p_{gy}, v_{pref}, \theta]$, $s^h \in R^4$. The state, position, and radius of the obstacles are described by $\hat{s}$, $\hat{p}$ and $\hat{r}$ respectively.

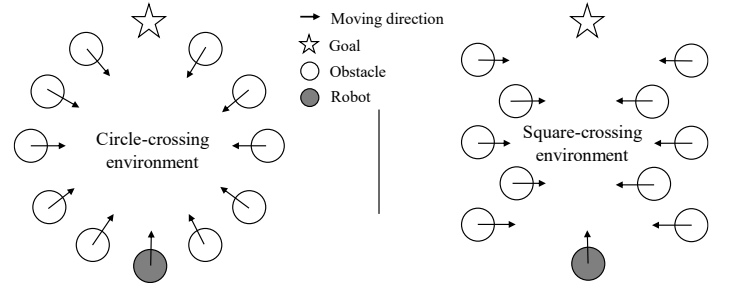

**Fig. 2.** Circle-crossing and square-crossing simulators. Obstacles are randomly generated near the brink of the circle in a circle-crossing environment. Then they move toward their opposite side. In the square-crossing environment, obstacles are randomly generated on the left side or right side and then they move toward random positions on their opposite side.

We first introduce one-robot one-obstacle case, and then the one-robot multi-obstacle case. The robot plans its motion by the policy $\pi: (s_{0:t}, \hat{s}_{0:t}^{obs}) \to a_t$ where $s_{0:t}$ and $\hat{s}_{0:t}^{obs}$ are the robot states and observable obstacle states from time step 0 to time step $t$, while the obstacles plan their motions by $\hat{\pi}: (\hat{s}_{0:t}, s_{0:t}^{obs}) \to a_t$ where $\hat{s}_{0:t}$ and $s_{0:t}^{obs}$ are the obstacle states and observable robot states from time step 0 to time step $t$. The robot's objective is to minimize the expectation (average) of the time to its goal $E[t_g]$ (1) under the policy $\pi$ without collisions to the obstacles. The constraints of robot's motion planning can be formulated via (2-5) that represent the *collision avoidance constraint*, *goal constraint*, *kinematics of the robot* and *kinematics of the obstacle*, respectively. Collision avoidance constraint denotes that the distance of robot and obstacles $\|p_t - \hat{p}_t\|_2$ should be greater than or equal to the radius sum of the robot and obstacles $r + \hat{r}$. Goal constraint denotes that the position of the robot $p_{tg}$ should be equal to the goal position $p_g$ if the robot reaches its goal. Kinematics of the robot denotes that the robot position $p_t$ is equal to the sum of robot position $p_{t-1}$ and the change of the robot position $\Delta t \cdot \pi: (s_{0:t}, \hat{s}_{0:t}^{obs})$. The robot policy $\pi: (s_{0:t}, \hat{s}_{0:t}^{obs})$ is a velocity decided by learning from historical robot states and obstacle states. Kinematics of the obstacle is the same as that of the robot. $\hat{\pi}: (\hat{s}_{0:t}^{obs}, s_{0:t})$ is a velocity decided by the obstacle policy $\hat{\pi}$ like ORCA.

$$minimize\, E[t_g | s_0, \hat{s}_0^{obs}, \pi, \hat{\pi}] \quad s.t. \quad (1)$$

$$\|p_t - \hat{p}_t\|_2 \geq r + \hat{r} \quad \forall t \quad (2)$$

$$p_{tg} = p_g \quad (3)$$

$$p_t = p_{t-1} + \Delta t \cdot \pi: (s_{0:t}, \hat{s}_{0:t}^{obs}) \quad (4)$$

$$\hat{p}_t = \hat{p}_{t-1} + \Delta t \cdot \hat{\pi}: (\hat{s}_{0:t}^{obs}, s_{0:t}) \quad (5)$$

In one-robot N-obstacle case, the objective is replaced by $minimize\, E[t_g | s_0, \{\hat{s}_0^{obs} ... \hat{s}_N^{obs}\}, \pi, \hat{\pi}]$ where we assume that obstacles use the same policy $\hat{\pi}$. Collision avoidance constraint is replaced by

$$\begin{cases} \|p_t - \hat{p}_{0:t}\|_2 \geq r + \hat{r} \\ \|p_t - \hat{p}_{1:t}\|_2 \geq r + \hat{r} \\ \quad \cdots \\ \|p_t - \hat{p}_{N-1:t}\|_2 \geq r + \hat{r} \end{cases} \forall t \quad (6)$$

assuming that obstacles are in the same radius $\hat{r}$. $\hat{p}_{N-1:t}$ denotes the position of *N*-th obstacle at the time step *t*. Kinematics of the robot is replaced by $p_t = p_{t-1} + \Delta t \cdot \pi: (s_{0:t}, \{s_{0:t}^{obs} \ldots s_{N-1:t}^{obs}\})$ where the historical states of all obstacles $\{s_{0:t}^{obs} \ldots s_{N-1:t}^{obs}\}$ are considered for generating the robot policy. Kinematics of the obstacles is replaced by

$$\begin{cases} \hat{p}_{0:t} = \hat{p}_{0:t-1} + \Delta t \cdot \hat{\pi} \\ \hat{p}_{1:t} = \hat{p}_{1:t-1} + \Delta t \cdot \hat{\pi} \\ \quad \cdots \\ \hat{p}_{N-1:t} = \hat{p}_{N-1:t-1} + \Delta t \cdot \hat{\pi} \end{cases} \quad (7)$$

*C. Preliminary of RL and discrete soft actor critic*

**Preliminary.** *Markov decision process* (MDP) is sequential decision process based on Markov Chain [31]. Markov Chain is defined by a variable set $X = \{X_n: n > 0\}$ where the probability $p(X_{t+1}|X_t, \ldots, X_1) = p(X_{t+1}|X_t)$. This means the state and action of the next step only depend on the state and action of the current step. MDP is described as a tuple $<S, A, P, R>$. *S* denotes the state and here it refers to the state of robot and obstacles. *A* denotes an action taken by the robot. Action $A = [\theta, v]$ is selected from *action space* where directions $\theta \in \{0, \frac{\pi}{8}, \ldots 2\pi\}$ and Speed of each direction $v \in \{0.2, 0.4, \ldots 1\}$. Hence, action space consists of 81 actions including a stop action. *P* denotes the possibility to transit from one state to the next state. *R* denotes the reward or punishment received by the robot after executing actions. The reward function in this paper is defined by

$$R(s, a) = \begin{cases} 1 & \text{if } p_{current} = p_g \\ -0.1 + \frac{d_{min}}{2} & \text{if } 0 < d_{min} < 0.2 \\ -0.25 & \text{if } d_{min} < 0 \\ \frac{d_{start\_to\_goal} - (p_g - p_{current})}{d_{start\_to\_goal}} \cdot 0.5 & \text{if } t = t_{max} \text{ and} \\ & p_t \neq p_g \\ 0 & \text{otherwise} \end{cases} \quad (8)$$

where $p_{current}$ denotes the position of the robot currently. $p_g$ denotes the position of the goal. $d_{min}$ denotes the minimum distance of the robot and obstacles during motion planning process. $d_{start\_to\_goal}$ denotes the distance of the start to the goal. $t_{max}$ is the allowed maximum time for any episode of the motion planning. Our reward function (8) is modified from [15] which cannot work without the imitation learning. (8) accelerates convergence speed by attaching a reward to *the final position of the robot*. This encourages the robot to approach the goal.

Other crucial terms of RL include the *value*, *policy*, *value function*, and *policy function*. Value denotes *how good one state is* or *how good one action is in one state*. The value consists of the *state value* (*V* value) and *state-action value* (*Q* value). Value is defined by the expectation of accumulative rewards $V(s) = \mathbb{E}[R_{t+1} + \gamma R_{t+1} + \cdots + \gamma^{T-1} R_T | s_t]$ or $Q(s, a) = \mathbb{E}[R_{t+1} + \gamma R_{t+1} + \cdots + \gamma^{T-1} R_T | (s_t, a_t)]$ where $\gamma$ is a discounted factor. The policy denotes the way to select actions. In function approximation case, policy is represented by the neural network.

Value function in deep RL scope is represented by neural networks to estimate the value of environmental state via the function approximation [32]. Policy function is also represented by neural networks. Actions are selected by indirect way (e.g., $a \leftarrow argmax_a R(s, a) + Q(s, a; \theta)$ in DQN [10][33]) or direct way (e.g., $\pi_\theta: s \rightarrow a$ in the actor-critic algorithm [34]).

**Discrete soft actor critic.** The policy of classical RL algorithm is obtained by maximizing the objective $\sum_{t=0}^{T} \mathbb{E}_{(s_t, a_t) \sim \rho_\pi}[r(s_t, a_t)]$. The objective of SAC is defined by the maximum entropy objective that considers the reward and entropy simultaneously

$$J(\pi) = \sum_{t=0}^{T} \mathbb{E}_{(s_t, a_t) \sim \rho_\pi}[r(s_t, a_t) + \alpha \mathcal{H}(\pi(\cdot|s_t))], \mathcal{H}(\pi(\cdot|s_t)) = -\log \pi(\cdot|s_t) \quad (9)$$

where $\mathcal{H}(\pi(\cdot|s_t))$ denotes the entropy. $\alpha$ is the temperature parameter. In objective maximization, SAC policy converges to optimal policy certainly by the soft policy iteration which consists of *policy evaluation* and *policy improvement*. Optimal policy is obtained by repeated application of policy evaluation and policy improvement. Policy evaluation [24] proves that if $Q^{k+1} = \mathcal{T}^\pi(Q^k)$, $Q^k$ will converge to the soft Q value of $\pi$ when $k \rightarrow \infty$. $\mathcal{T}^\pi(\cdot)$ is a modified Bellman backup operator given by

$$\mathcal{T}^\pi(Q)(s_t, a_t) \triangleq r(s_t, a_t) + \gamma \mathbb{E}_{s_{t+1} \sim p}[V(s_{t+1})] \quad (10)$$

where

$$V(s_t) = \mathbb{E}_{a_t \sim \pi}[Q(s_t, a_t) - \log \pi(a_t|s_t)]. \quad (11)$$

Applying $\mathcal{T}^\pi(\cdot)$ to Q value will bring Q value *closer* to $Q^\pi$. This means $Q(s_t, a_t) \leq \mathcal{T}^\pi(Q)(s_t, a_t) \leq Q^\pi(s_t, a_t)$. Policy improvement [24] proves that $Q^{\pi_{new}} \geq Q^{\pi_{old}}$ in objective maximization. $\pi_{new}$ is defined by

$$\pi_{new} = \arg\min_{\pi' \in \Pi} D_{KL}\left(\pi'(\cdot|s_t) \parallel \frac{\exp(Q^{\pi_{old}}(s_t, \cdot))}{Z^{\pi_{old}}(s_t)}\right) \quad (12)$$

where $Z^{\pi_{old}}(s_t)$ is the partition function for distribution normalization. It can be ignored because it does not contribute to the gradient of new policy. $Q^{\pi_{old}}$ guides the policy update to ensure an improved new policy. New policy is constrained to a parameterized family of distribution $\pi' \in \Pi$ like Gaussians to ensure the tractable and optimal new policy. Given the repeated application of policy evaluation and improvement, policy $\pi$ eventually converges to optimal policy $\pi^*$, $Q^{\pi^*} \geq Q^\pi, \pi \in \Pi$.

SAC is the combination of *soft policy iteration* and *function approximation*. In (9), temperature $\alpha$ is either a fixed value or an adaptive value. In function approximation, networks $\theta$, and $\phi$ are used to approximate the action value and policy value. The action value objective and its gradient are obtained by

$$\begin{cases} J(\theta) = \mathbb{E}_{(s_t, a_t) \sim \mathcal{D}}\left[\frac{1}{2}\left(Q(s_t, a_t; \theta) - \bar{Q}(s_t, a_t)\right)^2\right] \\ \bar{Q}(s_t, a_t) = r(s_t, a_t) + \gamma \mathbb{E}_{s_{t+1} \sim p}[V(s_{t+1}; \bar{\theta})] \\ \nabla_\theta J(\theta) \\ = \nabla_\theta Q(s_t, a_t; \theta) \cdot (Q(s_t, a_t; \theta) - r(s_t, a_t) + \gamma V(s_{t+1}; \bar{\theta}) \\ \quad - \alpha \log \pi_\phi(a_{t+1}|s_{t+1})) \end{cases} \quad (13)$$

where state value is approximated by $V(s_{t+1}; \bar{\theta})$. $\bar{\theta}$ is the target action value network. $\gamma$ is a discount factor. The policy objective and its gradient are obtained by

$$\begin{cases} J(\phi) = \mathbb{E}_{s_t \sim \mathcal{D}} \left[ D_{KL} \left( \pi_\phi(\cdot|s_t) \parallel \frac{\exp(Q(s_t,\cdot;\theta))}{Z_\theta(s_t)} \right) \right] \\ = \mathbb{E}_{s_t \sim \mathcal{D}}[\mathbb{E}_{a_t \sim \pi_\phi}[\alpha \log \pi_\phi(a_t|s_t) - Q(s_t, a_t; \theta)]] \\ \nabla_\phi J(\phi) = \nabla_\phi \alpha \log \pi_\phi(a_t|s_t) + \\ \nabla_\phi f_\phi(\epsilon_t; s_t) \cdot (\nabla_{a_t} \alpha \log \pi_\phi(a_t|s_t) - \nabla_{a_t} Q(s_t, a_t)) \\ a_t = f_\phi(\epsilon_t; s_t) \end{cases} \quad (14)$$

where $f_\phi(\epsilon_t; s_t)$ is the network transformation. $\epsilon_t$ is an input noise vector sampled from fixed distribution like spherical Gaussian. The temperature objective is defined by

$$J(\alpha) = \mathbb{E}_{a_t \sim \pi_t}[-\alpha \log \pi_t(a_t|s_t) - \alpha \bar{\mathcal{H}}] \quad (15)$$

where $\bar{\mathcal{H}}$ is the target entropy. Temperature objective gradient is obtained by approximating dual gradient descent [35]. Eventually, the networks and temperature are updated by

$$\begin{cases} \theta \leftarrow \theta - \gamma_\theta \nabla_\theta J(\theta) \\ \phi \leftarrow \phi - \gamma_\phi \nabla_\phi J(\phi) \\ \alpha \leftarrow \alpha - \gamma_\alpha \nabla_\alpha J(\alpha) \\ \bar{\theta} \leftarrow \tau \theta + (1 - \tau) \bar{\theta} \end{cases} \quad (16)$$

SAC is used in tasks with continuous action space. However, the action space in this paper is discrete. Hence, SAC should be modified to suit our task. Some modifications [25] should be made. They are summarized as the follows:

1) $Q$ function should be moved from $Q: S \times A \to \mathbb{R}$ to

$$Q: S \times A \to \mathbb{R}^{|A|}. \quad (17)$$

This means $Q$ values of all possible actions should be outputted, instead of a $Q$ value of the action taken by the robot.

2) The outputted policy should be the action distribution

$$\pi: S \to [0,1]^{|A|} \quad (18)$$

instead of the *mean* and *covariance* of action distribution of SAC $\pi: S \to \mathbb{R}^{2|A|}$.

3) In temperature objective (15), its expectation $\mathbb{E}_{a_t \sim \pi_t}[\cdot]$ is obtained by the Monte-Carlo estimation which involves taking an expectation over the action distribution [25]. In the discrete action space case, the expectation should be calculated directly, instead of Monte-Carlo estimation. Hence, the temperature objective changes into

$$J(\alpha) = \pi(s_t)^T [-\alpha \log \pi_t(s_t) - \alpha \bar{\mathcal{H}}] \quad (19)$$

Where $\bar{\mathcal{H}}$ is the target entropy. Similarly, the policy objective changes into

$$J(\phi) = \mathbb{E}_{s_t \sim \mathcal{D}}[\pi(s_t)^T [\alpha \log \pi_\phi(s_t) - Q(s_t; \theta)]] \quad (20)$$

III. METHOD

This section presents the implementations and optimizations of our motion planning algorithms. We first presents the implementation of the relational graph based DSAC. Then, relational graph based DSAC is improved by introducing the attention weight based DSAC. Finally, the attention weight based DSAC is further improved by integrating the skip connection method and LSTM pooling into the architecture of attention network of attention weight based DSAC.

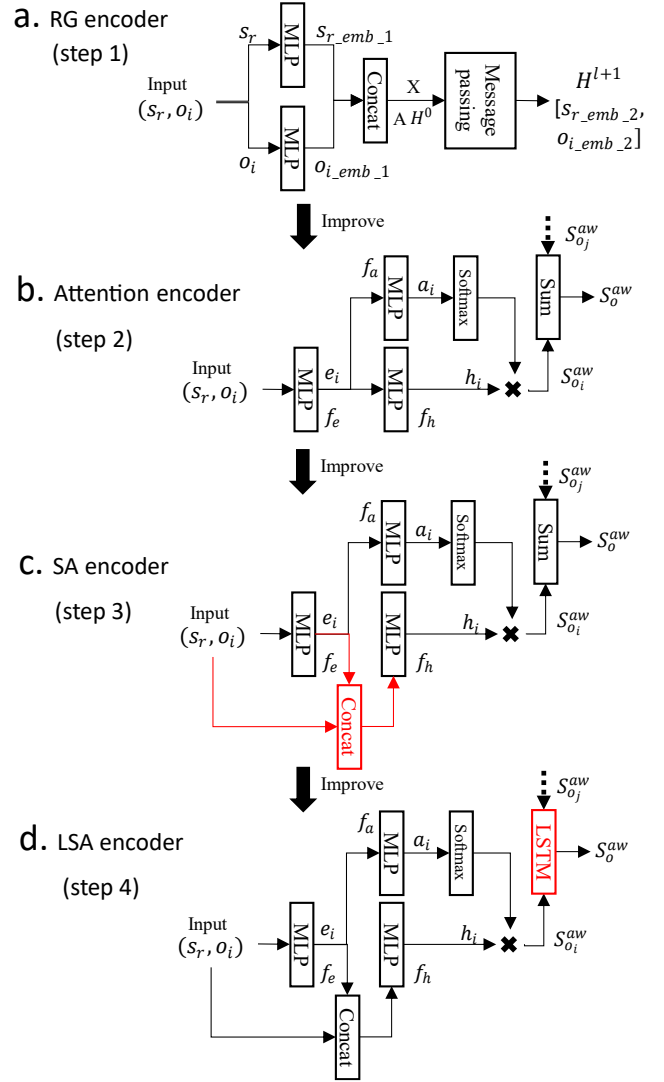

**Fig. 3.** Mechanisms of relational graph (a), attention weight (b), skip connection method for attention weight (c), and LSTM pooling and skip connection method for attention weight (LSA) (d). In the LSA encoder, the skip connection is for reducing the feature loss, while LSTM replaces the sum operator of the attention network to make the final interpreted features (environmental state) injective.

*A. Relation graph based DSAC (RG-DSAC)*

The mechanism of relational graph [13] is shown in Fig. 3a. The source input collected from the environment consists of source robot feature $s_r$ and $N$-obstacle features ($o_i, i \in 1,2,..N$). These features are in different dimensions. They should be formatted to the same dimension to accommodate the input requirement of the graph by

$$s_{r\_emb\_1} = MLP(s_r) \quad (21)$$

$$o_{i\_emb\_1} = MLP(o_i) \quad (22)$$

They are achieved by the *multi-layer perceptron* (MLP). $s_{r\_emb\_1}$ and $o_{i\_emb\_1}$ denote the embeddings of the robot and $i$-th obstacle feature. All obstacle features are concatenated to form the embedded obstacle state $s_{o\_emb\_1}$. The embedded robot state and embedded obstacle state are concatenated to form the initial feature matrix (environmental representation) $X$ by

$$X = concat(s_{r\_emb\_1}, s_{o\_emb\_1}) \quad (23)$$

The first row of $X$ ($X[0,:]$) denotes the robot feature and remaining rows ($X[i,:], i \in 1,2,..N$) denotes all obstacle features.

Given feature matrix $X$, relation matrix $A$ that represents the relationship of the *robot-obstacle and obstacle-obstacle* is computed by a *similarity function*. This is achieved by concatenating one of features in $X[i,:], i \in 0,1,..N$ with all features in $X[j,:], j \in 0,1,..N$ recursively to form the *pairwise features of the robot-obstacle and obstacle-obstacle*. Then, the relation feature $A[i,:]$ is obtained by MLP that maps these pairwise features to a fixed dimension (the same dimension as that of $X[i,:]$) via

$$A[i,:] = MLP\big(concat(X[i,:], X[:,:])\big). i \in 0,1,..N \quad (24)$$

Given feature matrix and relation matrix, interaction feature matrix $H$ is obtained by the *message passing rule* from the *graph convolutional network* (GCN) via

$$H^{l+1} = \sigma(AH^l W^l) + H^l, H^0 = H^1 = X, l \in 0,1.. \quad (25)$$

where $\sigma$ is the activation function. $W$ is layer-specific trainable weight, and $l$ denotes the number of the neural network layers. The *difference* between the initial feature matrix $X$ and the interaction feature matrix $H$ is that $H$ includes both initial features and relation features, while $X$ only includes initial features. Interaction feature matrix $H^{l+1}$ outperforms initial feature matrix $X$ in the *expressive power*. This is achieved via the relation matrix $A$ and the message passing, and it is shown by training and evaluation performances in the simulation and real-world motion planning. Moreover, the expressive power of the interaction feature can be further improved by LSTM pooling that maps interaction feature of obstacles $H^{l+1}[1:,:]$ to the sequential hidden features. Hence, final output of the relational graph (environmental state) that feeds DSAC consists of interaction feature of the robot and obtained sequential hidden features via

$$S^{rg} = [H^{l+1}[0,:], LSTM(H^{l+1}[1:,:])] \quad (26)$$

### B. *Attention weight based DSAC* (AW-DSAC)

Given the relational graph, it is obvious that relation matrix $A$ plays an essential role to improve the expressive power of output features. Relation matrix includes the robot-obstacle relation and obstacle-obstacle relation. Now, let's recall our task: robotic motion planning among dense obstacles. It is easy to see that robot-obstacle relation matters in our task. However, obstacle-obstacle relation does not show much *direct importance* in this task to generate features with high expressive power, although it has *marginal importance* to predict future obstacle trajectories that slightly improve the motion planning performance [13]. To further improve the motion planning performance, much attention should be paid to making the best of the robot-obstacle relation. Moreover, the importance of the obstacles also vary in different time step. The importance is shown by the robot speed, moving directions, and distance of robot and obstacle.

Recent attention weight mechanism [15] focuses on the pairwise robot-obstacle feature. It computes an attention score that weighs the importance of dynamic obstacles and makes the expressive power of interpreted environmental state interpretable. Hence, we apply the attention weight to replace relational graph for high and interpretable expressive power of the output features.

As the relational graph, in the attention weight case (Fig. 3b), the environmental state to feed DSAC $S^{aw}$ is defined by the feature combination of the robot and obstacles via

$$S^{aw} = [s_r, S_o^{aw}] \quad (27)$$

where $S_o^{aw}$ denotes the weighted obstacle feature, and it is defined by

$$S_o^{aw} = \sum_{i=1}^n [softmax(\alpha_i)] \cdot h_i \quad (28)$$

where $\alpha_i$ and $h_i$ denote the *attention score* and the *interaction feature* of the robot and the obstacle $o_i$ respectively. The interaction feature is a high-level feature that better outlines a robot-obstacle relation, compared to a shallow feature $e_i$. The interaction feature is defined by

$$h_i = f_h(e_i; w_h) \quad (29)$$

where $f_h(\cdot)$ and $w_h$ denote the MLP and its weight. $e_i$ denotes the *embedded shallow feature* obtained from the pairwise robot-obstacle feature $[s_r, o_i]$. The attention score is defined by

$$\alpha_i = f_\alpha(e_i; w_a) \quad (30)$$

where $f_\alpha(\cdot)$ and $w_a$ denote the MLP and its weight. The embedded shallow feature is defined by

$$e_i = f_e([s_r, o_i]; w_e), i \in 1,2,..N \quad (31)$$

where $f_e(\cdot)$ and $w_e$ denote the MLP and its weight.

### C. *Skip connection for Attention weight based DSAC* (SA-DSAC)

Recent progress in supervised DL [36][37][38] unveils that low-level (shallow) and high-level (deep) features play different role in the learning of the neural networks. The low-level feature provides more details of the source environmental state, while high-level feature outlines an overall structure of source environmental feature. Both low-level and high-level features contribute to the expressive power. Obviously, the attention weight mechanism just includes high-level feature to form the interaction feature $h_i$, given the mechanism of the attention weight in Fig 3b. This causes the loss of the details in environmental state and low expressive power of final feature $S^{aw}$ follows. To improve the expressive power of environmental state interpreted by attention weight, we introduce SA-DSAC that integrates the skip connection method (Fig. 3c) into the architecture of the attention network for generating optimized interaction feature by

$$h_i = f_h(concat(e_i, [s_r, o_i]); w_h) \quad (32)$$

where $f_h(\cdot)$ and $w_h$ denote the MLP and its weight. $e_i$ denotes the *embedded shallow feature* obtained from the pairwise robot-obstacle feature $[s_r, o_i]$.

### D. *LSTM pooling and Skip connection for Attention weight based DSAC* (LSA-DSAC)

Given the attention weight mechanism, we can notice that weighted obstacle features $S_o^{aw}$ are pooled by summing all weighted interaction features. Recent research [39] unveils a high performance of the sum operation over the *mean* and *max* operations in pooling features for generating new features with high expressive power. However, it does not mean that sum operation is absolute *injective* [39]. The more injective the feature is, the more distinguishable the feature is against other features.

Hence, high injectivity of feature means high expressive power [39]. Sum operation just outlines an overall structure of pooled features, and some pooled features based on the sum operation lack injectivity or are undistinguishable. For instance, $sum$ (3,1) and $sum$ (2,2). Features [3,1] and [2,2] are equal statistically, but they are obviously different features. We think that *keeping some source features in the feature pooling process* contributes to the injectivity of pooled features.

LSTM pooling is expected to be a good solution to achieve this goal where the source features are just mapped into sequential hidden features. In this process, the structural information, and a part of the feature details of the source features are kept, instead of just keeping the statistical property of source features via the sum operation. Hence, we introduce LSA-DSAC that takes LSTM to replace sum operation in the pooling of weighted obstacle feature $S_o^{aw}$. LSTM maps weighted interaction features $softmax(\alpha_i) \cdot h_i$ to sequential features (Fig. 3d). This better preserve the feature of each weighted interaction feature. This is achieved by

$$S_o^{aw} = LSTM[softmax(\alpha_i) \cdot h_i], i \in 1,2,...N \quad (33)$$

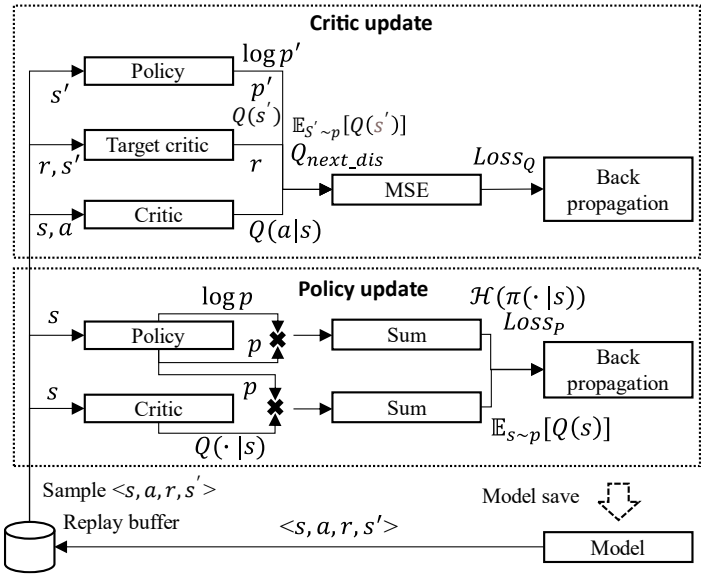

**Fig. 4.** Training process of our LSA-DSAC. LSA-DSAC starts by collecting data from the environment using the initialized models. Collected data is saved in the replay buffer from which data is sampled. The policy and critic are updated based on sampled data until convergence, resulting in trained models which are saved for evaluations in the motion planning tasks.

Once the environmental state $S^{aw} = [s_r, S_o^{aw}]$ generated by LSA is well prepared, it will feed DSAC to generate trained models for the motion planning of the robot. In implementation, separate attention networks are taken to form the *critic* and *policy* of DSAC by

$$critic = [\theta_{att_c}, \theta_{c1}, \theta_{c2}] \quad (34)$$

$$policy = [\theta_{att\_p}, \theta_p] \quad (35)$$

where double-network architecture (networks $\theta_{c1}, \theta_{c2}$) is taken in critic to reduce the overestimation of the Q value, while policy just has single network $\theta_p$ for prediction. The attention network connects with the prediction network to form the critic or policy of DSAC (Alg. 1). The training process of LSA-DSAC is shown in Fig. 4. Episodic data $<s,a,r,s'>$ of each time step is obtained by performing the policy of DSAC

$$<s_t, a_t, r_t, S_{t+1}> \sim \pi(a_t|s_t; [\theta_{att\_p}, \theta_p]) \quad (36)$$

Episodic data is stored in replay buffer $\mathcal{D}$ at the end of each episode by

$$\mathcal{D} \leftarrow \mathcal{D} \cup \mathcal{E}, \mathcal{E}=\mathcal{E}+<s_t, a_t, r_t, S_{t+1}> \quad (37)$$

Networks are trained in each step of an episode (Alg. 2). In the forward propagation process, the critic loss $Loss_Q$ and policy loss $Loss_P$ are obtained by

$$Loss_Q = MSE(Q(a|s)_{c1}, Q_{next\_dis}) + MSE(Q(a|s)_{c2}, Q_{next\_dis}) \quad (38)$$

$$Loss_P = -mean[\mathbb{E}_{s\sim p}[Q(s)] + \alpha \cdot \mathcal{H}(\pi(\cdot|s))] \quad (39)$$

where $Q_{next\_dis}$ and $Q(a|s)_{ci}, i \in \{1,2\}$ denote the discounted next state value and current action values respectively. $\mathbb{E}_{s\sim p}[Q(s)]$ and $\mathcal{H}(\pi(\cdot|s))$ denote the expectation of current state value and current policy entropy respectively. $\alpha$ denotes the temperature parameter.

**Compute discounted next state value.** $Q_{next\_dis}$ is computed by

$$Q_{next\_dis} = r(s,a) + \gamma \mathbb{E}_{S'\sim p}[Q(s')] \quad (40)$$

where $r(s,a)$ denotes the reward from the environment after executing action $a$ in state $s$. $\mathbb{E}_{S'\sim p}[Q(s')]$ denotes the expectation of the next state value. $\gamma$ denotes a discount factor. $\mathbb{E}_{S'\sim p}[Q(s')]$ is computed by

$$\mathbb{E}_{S'\sim p}[Q(s')] = \sum[p' \cdot \min(Q(s')_{c1}, Q(s')_{c2}) - \alpha \cdot \log p'] \quad (41)$$

where $Q(s')_{c1}$ and $Q(s')_{c2}$ denote the next state values computed by the target critic via the algorithm *Forward-propagation-critic*. $p'$ and $\log p'$ denote the next policy distribution and its logit value. They are computed by *Forward-propagation-policy*.

The forward propagations of the critic and policy (Alg. 3-4) are almost the same. Their difference is that the critic takes two networks to compute two Q values by

$$Q(s)_{c1}, Q(s)_{c2} \leftarrow f_{\theta_{ci}}(S^{aw}), i \in 1,2 \quad (42)$$

Then, an average Q value is obtained. This reduces the bias (overestimation) of the Q value. The policy just takes single network to compute policy distribution and its logit value by

$$\log p, p \leftarrow f_{\theta_p}(S^{aw}) \quad (43)$$

**Compute current action values.** To obtain $Q(a|s)_{ci}, i \in \{1,2\}$, current state values $Q(s)_{ci}, i \in \{1,2\}$ should be computed first by the algorithm *Forward-propagation-critic*. Then, current action values are computed by gathering state value along the policy distribution of the action $a$ via

$$Q(a|s)_{c1} = Q(s)_{c1}.gather(a), Q(a|s)_{c2} = Q(s)_{c2}.gather(a) \quad (44)$$

**Compute expectation of current state value.** The process to compute the expectation of current state value is different from that of the next state value. It is computed by

$$\mathbb{E}_{s\sim p}[Q(s)] = \sum \min[(Q(s)_{c1}, Q(s)_{c2}) \cdot p] \quad (45)$$

where $Q(s)_{ci}, i \in \{1,2\}$ and $p$ are computed respectively by *Forward-propagation-critic* and *Forward-propagation-policy*.

**Compute policy entropy.** $\mathcal{H}(\pi(\cdot|s))$ is computed by

$$\mathcal{H}(\pi(\cdot|s)) = -\log \pi(\cdot|s) = -\sum p \cdot \log p \quad (46)$$

Before the back-propagation process, the temperature loss $\mathcal{L}_\alpha$ is also required for the network update. $\mathcal{L}_\alpha$ is computed by

$$\mathcal{L}_\alpha = -\min[\log \alpha \cdot (\bar{\mathcal{H}} - \mathcal{H})] \quad (47)$$

where $\bar{\mathcal{H}}$ is the target entropy. Then, the temperature and all networks are updated by the gradient ascent via

$$\theta_{att\_c,ci} \leftarrow \theta_{att\_c,ci} - \gamma \nabla_{\theta_{att\_c,ci}} \mathcal{L}oss_Q, i \in 1,2 \quad (48)$$

$$\theta_{att\_p,p} \leftarrow \theta_{att\_p,p} - \gamma \nabla_{\theta_{att\_p,p}} \mathcal{L}oss_P \quad (49)$$

$$\alpha \leftarrow \alpha - \gamma \nabla_\alpha \mathcal{L}_\alpha, \; \alpha \leftarrow e^\alpha \quad (50)$$

Finally, target critic is also updated for a new training round via

$$\bar{\theta}_{att\_c} \leftarrow \theta_{att\_c}, \bar{\theta}_{c1} \leftarrow \theta_{c1}, \bar{\theta}_{c2} \leftarrow \theta_{c2} \quad (51)$$

**Algorithm 1: LSA-DSAC**

1. Initialize the replay buffer $\mathcal{D}$
2. Initialize attention net of critic $\theta_{att\_c}$, attention net of policy $\theta_{att\_p}$, prediction nets of critic $\theta_{c1}$ and $\theta_{c2}$, and prediction net of policy $\theta_p$ where
$$critic = [\theta_{att_c}, \theta_{c1}, \theta_{c2}], policy = [\theta_{att_p}, \theta_p]$$
3. Initialize target critic $[\bar{\theta}_{att\_c}, \bar{\theta}_{c1}, \bar{\theta}_{c2}]$:
$$\bar{\theta}_{att\_c} \leftarrow \theta_{att\_c}, \bar{\theta}_{c1} \leftarrow \theta_{c1}, \bar{\theta}_{c2} \leftarrow \theta_{c2}$$
4. **For** episode $i < N$ **do**
5.     **For** $t \neq T_{terminal}$ in episode $i$ **do**
6.         Execute action:
$$<s_t, a_t, r_t, S_{t+1}> \sim \pi(a_t|s_t; [\theta_{att\_p}, \theta_p])$$
7.         **Train** If length $(\mathcal{D})$ < batch size $l$
8.         Store data of this episode:
$$\mathcal{E} = \mathcal{E} + <s_t, a_t, r_t, S_{t+1}>$$
9.     Update replay buffer: $\mathcal{D} \leftarrow \mathcal{D} \cup \mathcal{E}$
10.    $i = i + 1$
11. Save models: $\theta_{att\_c}, \theta_{att\_p}, \theta_{c1}, \theta_{c2}$ and $\theta_p$

**Algorithm 2: Train**

1. Sample $K$-batch experiences randomly from replay buffer $\mathcal{D}$

//**Prepare discounted next state value $Q_{next\_dis}$**

2. Compute next policy distribution $p'$ and its logit value $\log p'$:
    **Forward-propagation-policy**
3. Compute next state value $Q(s')_{c1}, Q(s')_{c2}$ by target critic:
    **Forward-propagation-critic**
4. Compute expectation of next state value:
$$\mathbb{E}_{S' \sim p}[Q(s')] = \sum [p' \cdot \min(Q(s')_{c1}, Q(s')_{c2}) - \alpha \cdot \log p']$$
5. Compute discounted next state value:
$$Q_{next\_dis} = r(s,a) + \gamma \mathbb{E}_{S' \sim p}[Q(s')]$$

//**Prepare current action value $Q(a|s)_{c1}, Q(a|s)_{c2}$**

6. Compute current state value $Q(s)_{c1}, Q(s)_{c2}$:
    **Forward-propagation-critic**
7. Compute current action value $Q(a|s)_{c1}, Q(a|s)_{c2}$:
$$Q(a|s)_{c1} = Q(s)_{c1}.gather(a)$$
$$Q(a|s)_{c2} = Q(s)_{c2}.gather(a)$$

//**Prepare Q value loss (critic loss)**

8. Compute Q value loss:
$$\mathcal{L}oss_Q = MSE(Q(a|s)_{c1}, Q_{next\_dis}) + MSE(Q(a|s)_{c2}, Q_{next\_dis})$$

//**Prepare policy entropy**

9. Compute policy distribution $p$ and its logit value $\log p$:
    **Forward-propagation-policy**
10. Compute current policy entropy:
$$\mathcal{H}(\pi(\cdot|s)) = -\log \pi(\cdot|s) = -\sum p \cdot \log p$$

//**Prepare expectation of current state value**

11. Compute current state value $Q(\cdot|s)_{c1}, Q(\cdot|s)_{c2}$ for all actions where $Q(\cdot|s) = Q(s)$:
    **Forward-propagation-critic**
12. Compute expectation of current state value:
$$\mathbb{E}_{s \sim p}[Q(s)] = \sum \min[(Q(s)_{c1}, Q(s)_{c2}) \cdot p]$$

//**Prepare policy and temperature loss**

13. Compute policy loss:
$$\mathcal{L}oss_P = -mean[\mathbb{E}_{s \sim p}[Q(s)] + \alpha \cdot \mathcal{H}(\pi(\cdot|s))]$$
14. Compute temperature loss:
$$\mathcal{L}_\alpha = -\min[\log \alpha \cdot (\bar{\mathcal{H}} - \mathcal{H})]$$

//**Back-propagation to update networks and temperature**

15. Update attention net $\theta_{att\_c}$ and prediction nets of critic $\theta_{c1}$ and $\theta_{c2}$:
$$\theta_{att\_c,ci} \leftarrow \theta_{att\_c,ci} - \gamma \nabla_{\theta_{att\_c,ci}} \mathcal{L}oss_Q, i \in 1,2$$
16. Update attention net $\theta_{att\_p}$ and prediction nets of policy $\theta_p$:
$$\theta_{att\_p,p} \leftarrow \theta_{att\_p,p} - \gamma \nabla_{\theta_{att\_p,p}} \mathcal{L}oss_P$$
17. Update temperature:
$$\alpha \leftarrow \alpha - \gamma \nabla_\alpha \mathcal{L}_\alpha, \alpha \leftarrow e^\alpha$$
18. Update target critic $[\bar{\theta}_{att\_c}, \bar{\theta}_{c1}, \bar{\theta}_{c2}]$:
$$\bar{\theta}_{att\_c} \leftarrow \theta_{att\_c}, \bar{\theta}_{c1} \leftarrow \theta_{c1}, \bar{\theta}_{c2} \leftarrow \theta_{c2}$$

**Algorithm 3: Forward-propagation-critic**

//**Prepare input to feed prediction nets $\theta_{ci}, i \in \{1,2\}$**

1. Calculate embedded shallow feature:
$$e_i = f_e(s; w_e), i \in 1,2,..N$$
2. Calculate optimized interaction feature:
$$h_i = f_h(concat(e_i, [s_r, o_i]); w_h)$$
3. Calculate attention score and its softmax value:
$$\alpha_i = f_\alpha(e_i; w_a)$$
4. Calculate weighted feature of each obstacle:
$$[softmax(\alpha_i)] \cdot h_i, i \in 1,2,..N$$
5. Pool weighted feature of obstacles by LSTM:
$$S_o^{aw} = LSTM[softmax(\alpha_i) \cdot h_i], i \in 1,2,..N$$
6. Obtain input data $S^{aw}$ for prediction net: $S^{aw} = [s_r, S_o^{aw}]$

//**Prepare state value of $\theta_{c1}$ and $\theta_{c2}$**

7. Calculate state values: $Q(s)_{c1}, Q(s)_{c2} \leftarrow f_{\theta_{ci}}(S^{aw})$

**Algorithm 4: Forward-propagation-policy**

//**Prepare input to feed prediction net $\theta_p$**

1. Calculate shallow feature: $e_i = f_e(s; w_e), i \in 1,2,..N$
2. Calculate optimized interaction feature:
$$h_i = f_h(concat(e_i, [s_r, o_i]); w_h)$$
3. Calculate attention score and its softmax value:
$$\alpha_i = f_\alpha(e_i; w_a)$$
4. Calculate weighted feature of each obstacle:
$$[softmax(\alpha_i)] \cdot h_i, i \in 1,2,..N$$
5. Pool weighted feature of obstacles by LSTM:
$$S_o^{aw} = LSTM[softmax(\alpha_i) \cdot h_i], i \in 1,2,..N$$
6. Obtain input data $S^{aw}$ for prediction net: $S^{aw} = [s_r, S_o^{aw}]$

//**Prepare policy distribution and its logit value**

7. Calculate policy distribution and its logit value:
$$\log p, p \leftarrow f_{\theta_p}(S^{aw})$$

## IV. EXPERIMENTS

This section presents the implementation details of our algorithms. First, the network framework of our LSA-DSAC is given. Second, the model training details of our algorithms and the state-of-the-art are presented. Third, the model evaluations are conducted. The evaluations include the converged reward evaluation, interpretability evaluation, qualitative evaluation, quantitative evaluation, time complexity evaluation, transferability evaluation, and robustness evaluation. Finally, the physical implementation is presented.

### A. Network framework

In implementation, the network framework of our LSA-DSAC (Fig. 5) takes the architecture with separate attention networks. The prediction networks of the critic and policy connect with different attention network to form the critic and policy of DSAC. This contributes to overall convergence, compared with architecture with a shared attention network. The prediction network of the critic consists of two linear networks. Each linear network has three linear layers. The prediction network of the policy just has one linear network that also has three linear layers.

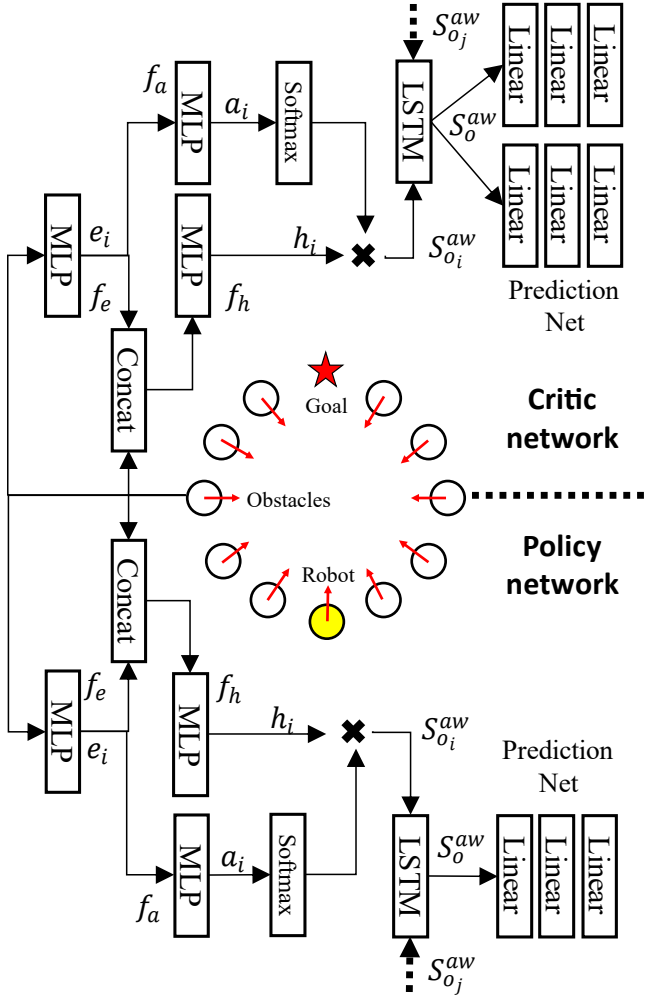

**Fig. 5.** Network framework of our LSA-DSAC. The framework of LSA-DSAC consists of the critic network and policy network. The critic network or policy network receives the same environment features from the circle-crossing simulator for the training. Both critic network and policy network consist of feature interpretation part and prediction part. The feature interpretation part is our LSA encoder which translates features of the robot and obstacles into the attention-based environment features $S^{aw}$. Attention-based environment features $S^{aw}$ is the combination of the robot feature $s_r$ and attention-based obstacle features $S_o^{aw}$. Attention-based environment features $S^{aw}$ are then fed into prediction networks which are the linear neural layers. Double prediction networks are used in the critic network to reduce the overestimation of Q value, while single prediction network is used in the policy network.

### B. Model training

We first implemented RG-DSAC for motion planning in the circle-crossing simulator. The training result (Fig. 6a) demonstrated that RG-DSAC converged faster than DSAC with source environmental state. Moreover, the converged result of RG-DSAC also outperformed the DSAC with source environmental state.

The output of the relational graph is the matrix $H^{l+1}$ (layer number $l$=2). The final interpreted feature to feed DSAC $S^{rg}$ is the combination of the robot feature $H^{l+1}[0,:]$ and the pooled obstacle features $LSTM(H^{l+1}[1:,:])$ where obstacle features are pooled by LSTM. To prove the efficacy and efficiency of our method to prepare the final features for training (*rob+lstm(obs)*), we compared it with other potential features for training by the ablation experiments where other features include:

1) features based on the feature concatenation of the robot and obstacles (*rob+obs*),
2) source robot feature (*rob*),
3) features from summing concatenated feature of the robot and obstacle (*sum(rob+obs)*),
4) features from concatenating the robot feature and obstacle features pooled by MLP (*rob+mlp(obs)*),
5) pooled features of the robot and obstacles by MLP (*mlp(rob+obs)*), and
6) pooled features of the robot and obstacles by LSTM (*lstm(rob+obs)*).

Ablation experiments (Fig. 6b) demonstrate that interpreted features for training using our method outperforms other potential features. Experiments also indicate that the robot feature should be separated from the obstacle features pooled by LSTM or MLP, resulting in the features *rob+lstm(obs)* and *rob+mlp(obs)*. This contributes to the expressive power of interpreted features for training. The interpreted feature *rob+mlp(obs)* marginally contributes to the convergence, while the interpreted feature *rob+lstm(obs)* by our LSA-DSAC dramatically improves the convergence.

We noticed that a separate architecture (two relation networks and two LSTM) outperforms a shared architecture (one shared relation network with shared and separate LSTMs) (Fig. 6c) in the implementation of RG-DSAC. Experiment shows that separate LSTM encoding contributes to the convergence dramatically (yellow and green curves), while a separate relation network also contributes to the convergence (blue and yellow curves).

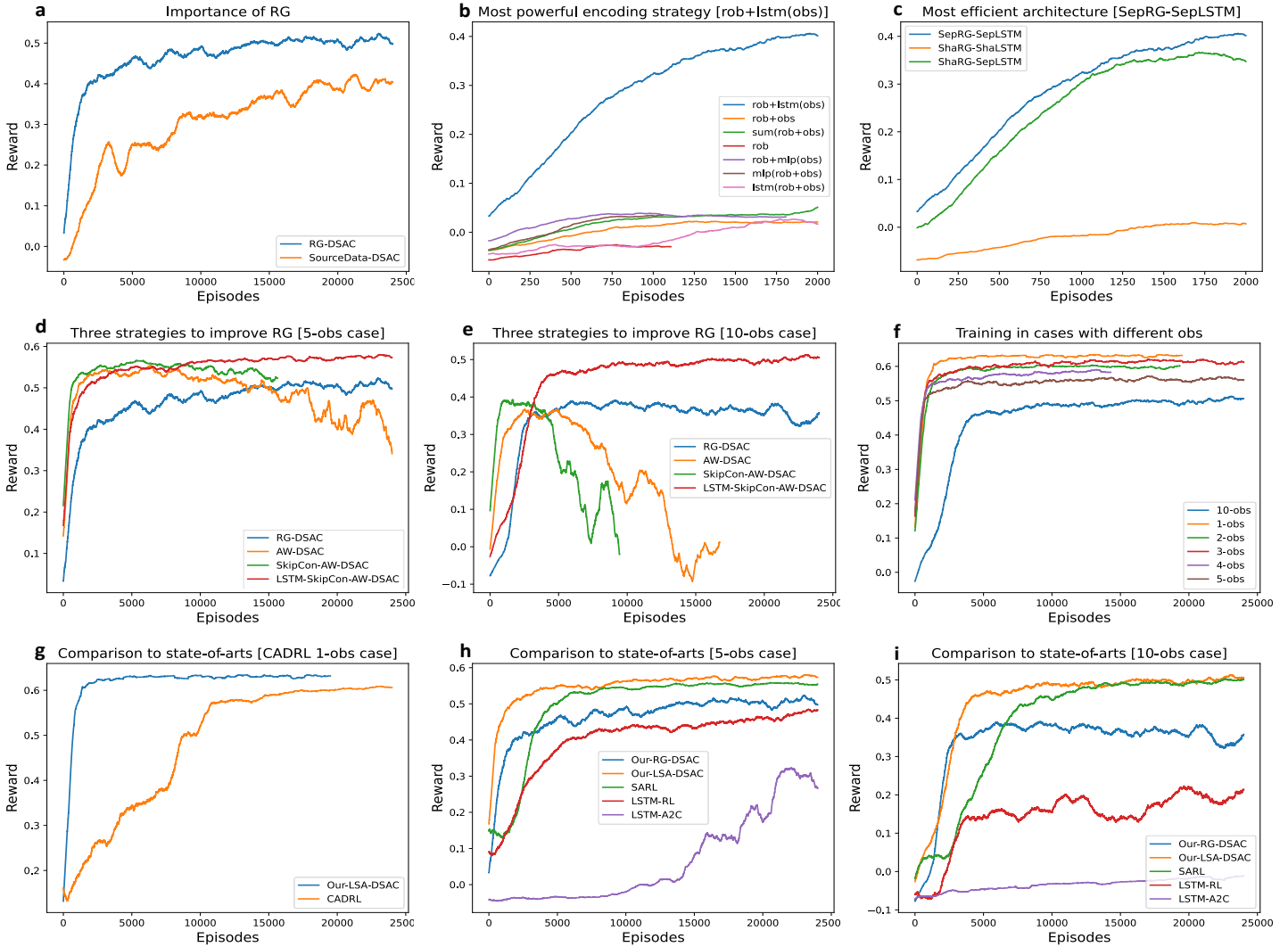

**Fig. 6.** The training curves of our algorithms and state-of-the-art. (a) denotes that the environment state interpreted by the relational graph contributes more to the convergence of DSAC than the source environment state. (b) denotes that after acquiring the robot and obstacle features interpreted by the relational graph, the obstacle features should be encoded by the LSTM. Then, the robot features concatenate the LSTM-encoded obstacle features to form new features for the learning of DSAC. This method of preparing new features outperforms the rest of the methods in improving the convergence of DSAC. (c) denotes that the network architecture with the separate relational graph and separate LSTM (Fig. 5) contributes more to the convergence of DSAC than the architecture with the shared relational graph or share LSTM. (d-e) denotes that the feature interpretation based on the relational graph partly represents the relationship between the robot and the obstacles, resulting in a slow convergence of DSAC. The attention weight or attention network focuses on and precisely describes the relationship between the robot and the obstacles, resulting in a fast convergence of DSAC. However, attention weight overfits in the training, and the overfitting problem can be mitigated by applying the skip connection method and LSTM pooling method. (f) denotes that the convergence of LSA-DSAC becomes slow, with the increase of the dynamic obstacles in the environment. (g) denotes the comparison of LSA-DSAC and CADRL in convergence. CADRL does not support multi-agent training, therefore the comparison of LSA-DSAC and CADRL is presented in a separate figure. (h-i) denotes the comparisons of LSA-DSAC, RG-DSAC, and the state-of-the-arts that support multi-agent training.

In Fig. 6d, the convergence is improved after the attention network replaces the relational graph for the feature interpretation. Then, the convergence is further improved by integrating the skip connection method and LSTM pooling to the attention network. Experiment also shows that AW-DSAC and SA-DSAC *overfit* in training because of the sum operation that lacks robustness or injectivity. LSA-DSAC outperforms the rest algorithms in converged result by sacrificing the convergence speed at early-stage training. The experiment is extended to 10-obstacle cases (Fig. 6e) where our LSA-DSAC still outperforms the rest algorithms in overall convergence speed and converged result. LSA-DSAC is also trained in cases with 1, 2, 3 and 4 obstacles (Fig. 6f). The experiment shows that the increase of the environmental complexity (the number of dynamic obstacles) results in a decrease of convergence.

Finally, LSA-DSAC is compared with RG-DSAC and the state-of-the-art that includes CADRL, LSTM-A2C, LSTM-RL and SARL. Note that ORCA is not trainable and it is not included in the training comparisons. LSA-DSAC is compared with CADRL in 1-obstacle case because CADRL only supports single-obstacle training (Fig. 6g). The rest state-of-the-art supports multi-obstacle training and they are trained in cases with 5 and 10 obstacles (Fig. 6h-i). Experiment shows that our LSA-DSAC is superior to the state-of-the-art in both convergence speed and converged result. The training parameters of our LSA-DSAC is shown in TABLE I.

TABLE I. Parameters and hyper-parameters of LSA-DSAC

| Parameters/ Hyper-parameters | Values of Parameters/ Hyper-parameters |
|---|---|
| LSTM hidden size | 50 |
| Number of MLP layer | 3 |
| ReLU layer after MLP | Yes (First MLP layer) |
| MLP input/output size (interaction and embedded layers) | [150, 100]-[100, 50] |
| MLP input/output size (attention layer) | [100, 100]-[100, 1] |
| Reward | Source reward |
| Gamma | 0.95 |
| Tau | 0.005 |
| Learning rate | 3e-4 |
| Alpha | 0.2 |
| Frequency of network update | Per step |
| Automatic entropy tuning | True |
| Batch size | 128 |
| Input layer size (DSAC network) | 6+50 |
| Hidden layer size (DSAC network) | 128 |
| Output size of policy network (DSAC network) | 81 |

*C. Model evaluation*

Trained models of all algorithms are evaluated in 5-obstacle case comprehensively in the circle-crossing simulator from seven perspectives. This includes the *converged reward evaluation*, *interpretability (explainable ability) evaluation*, *qualitative evaluation*, *quantitative evaluation*, *time complexity evaluation*, *transferability evaluation*, and *robustness evaluation*. Models are evaluated with 500 test sets (episodes).

**Converged reward evaluation.** Converged reward indicates the overall performance of the trained models in small test set during training. It provides a fast impression about how good the model is in training. It is easy to see that our LSA-DSAC outperforms the state-of-the-art in the converged reward, while SARL performs best among the state-of-the-art (TABLE II). CADRL just supports training with one obstacle, therefore its model is not included in comparison.

**Interpretability (explainable ability) evaluation.** Interpretability (explainable ability) here is defined as the ability to decide *directly, explicitly, and uniformly* how good the motion planning performance is. Attention mechanism (attention network) provides the attention score (a post-training indicator) to evaluate the importance of obstacles, therefore justifying the robot's policy or actions. Then, motion planning strategy is generated based on the attention score. Motion planning strategy (Fig. 7) of our LSA-DSAC indicates that the attention score is an overall evaluation that considers the moving direction, moving speed, and distance of robot and obstacles (e.g., humans). The distance of robot and obstacles sometimes contributes less to the attention score (e.g., human 2 in Fig. 7 that has minimum distance to the robot). Interpretability comparisons of the models are shown in TABLE III. Our LSA-DSAC and SARL have interpretability because of the attention score, while the motion planning performance of the rest algorithms cannot be observed indirectly, explicitly, and unexplainably.

TABLE II. Converged reward of the models in training with five dynamic obstacles. ORCA is not trainable, therefore it is not included in the comparisons. CADRL does not support multi-agent training, and the converged result of CADRL is from the training with one dynamic obstacle.

| Algorithms | Converged reward |
|---|---|
| ORCA [8] | --- |
| CADRL [27] | 0.61(training with one obstacle) |
| LSTM-RL [15][10] | 0.49 |
| SARL [15] | 0.55 |
| LSTM-A2C [29][11][28] | 0.30 |
| Our RG-DSAC | 0.50 |
| Our LSA-DSAC | **0.57** |

TABLE III. Interpretability (explainable ability) evaluation. Here the interpretability is measured by whether there are post-training indicators generated to justify the actions or policies.

| Algorithm | Interpretability (explainable ability) |
|---|---|
| ORCA | --- |
| CADRL | No |
| LSTMRL | No |
| SARL | Yes |
| LSTM-A2C | No |
| RG-DSAC | No |
| LSA-DSAC | **Yes** |

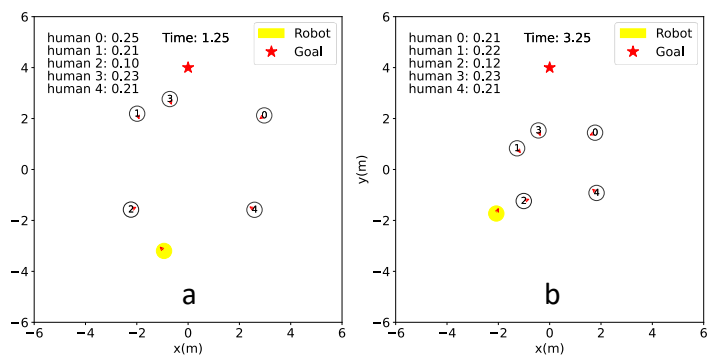

**Fig. 7.** Examples of obstacles with attention score in LSA-DSAC. The human here denotes the dynamic obstacles. The distance between the robot and obstacles is not so important to decide the attention score sometimes, such as the human (obstacle) 2 in (a) and (b). The obstacles heading toward the robot, such as human 0, are expected to have a higher attention score, but it doesn't mean only the direction of the obstacle decides the attention score. Attention score is an overall evaluation that considers the direction of motion, speed, and distance between the robot and the obstacles. Hence, in (b), human 0 has a high attention score, but its attention score is slightly smaller than that of human 1 and human 3.

TABLE IV. Features of six motion planning strategies. The motion planning strategy here is defined by humans according to human experience.

| Strategy | Description | Speed | Collision |
| --- | --- | --- | --- |
| Full-bypass | Bypass all obstacles | Fast | Less |
| Partly-bypass | Bypass most obstacles | Fast | Less |
| Follow-pass | Follow front obstacles and pass | Medium | Less |
| Wait-pass | Wait until obstacles move away and pass | Slow | Less |
| Back-pass | Move back until obstacles move away and pass | Slow | Less |
| Cross | Cross dense obstacles | High/Medium/Slow | More |

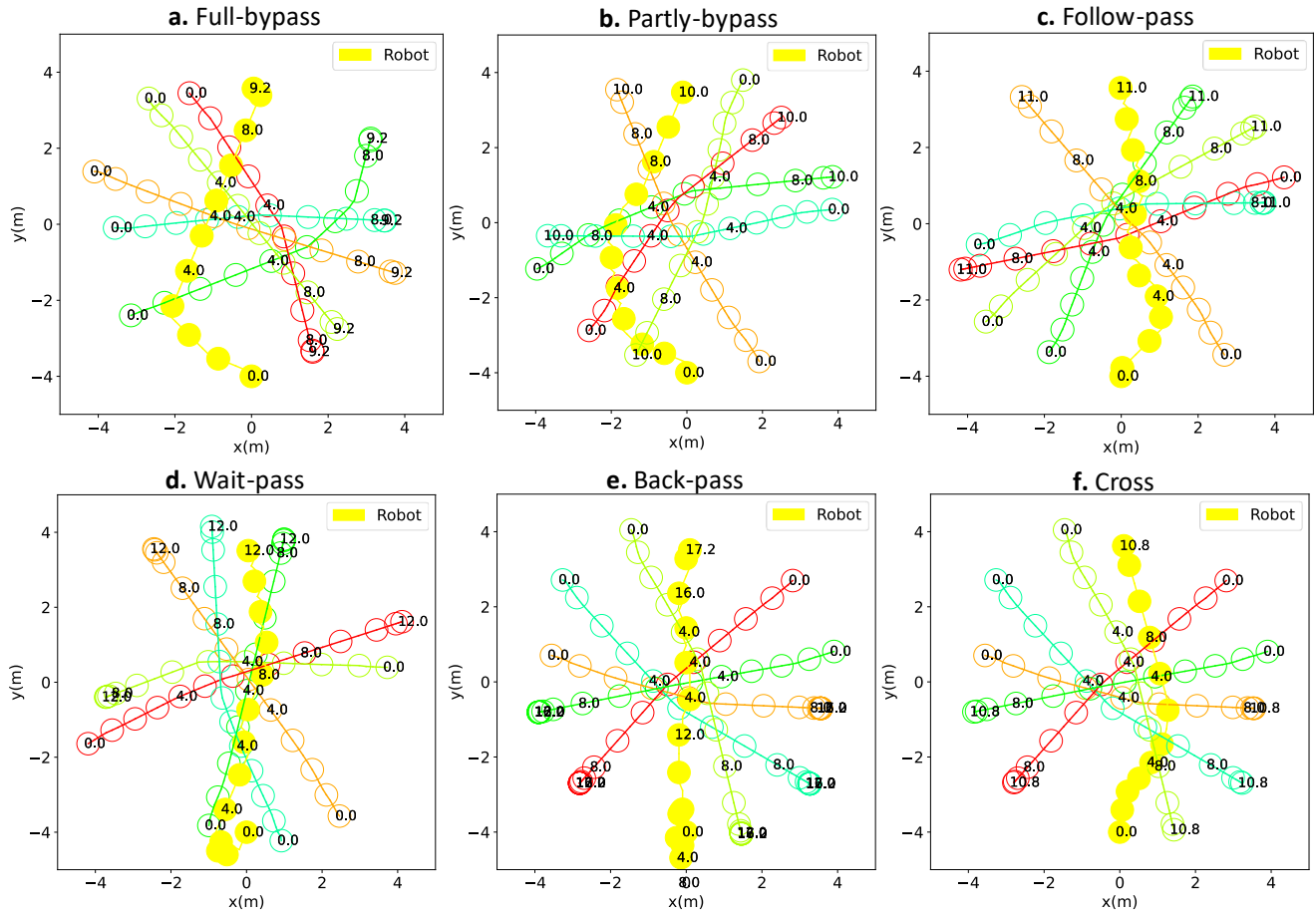

Fig. 8. Six learned motion planning strategies. The numbers along the trajectories represent the time step of each robot or obstacle. The full-bypass and partly-bypass are the most efficient motion planning strategies. The performance of wait-pass and follow-pass strategies is acceptable. The back-pass is the most time-consuming motion planning strategy. The cross strategy is efficient in the motion planning sometimes, but it causes many collisions.

TABLE V. Qualitative evaluation. The quality here is measured according to the efficiency or the property of learned motion planning strategies.

| Algorithm | Learnt motion planning strategy |
| --- | --- |
| ORCA | Cross |
| CADRL | Cross/Follow-pass |
| LSTMRL | Partly-bypass/Follow-pass |
| SARL | Full-bypass/Partly-bypass |
| LSTM-A2C | Back-pass/Wait-pass |
| RG-DSAC | Partly-bypass/Follow-pass |
| LSA-DSAC | **Full-bypass/**Partly-bypass |

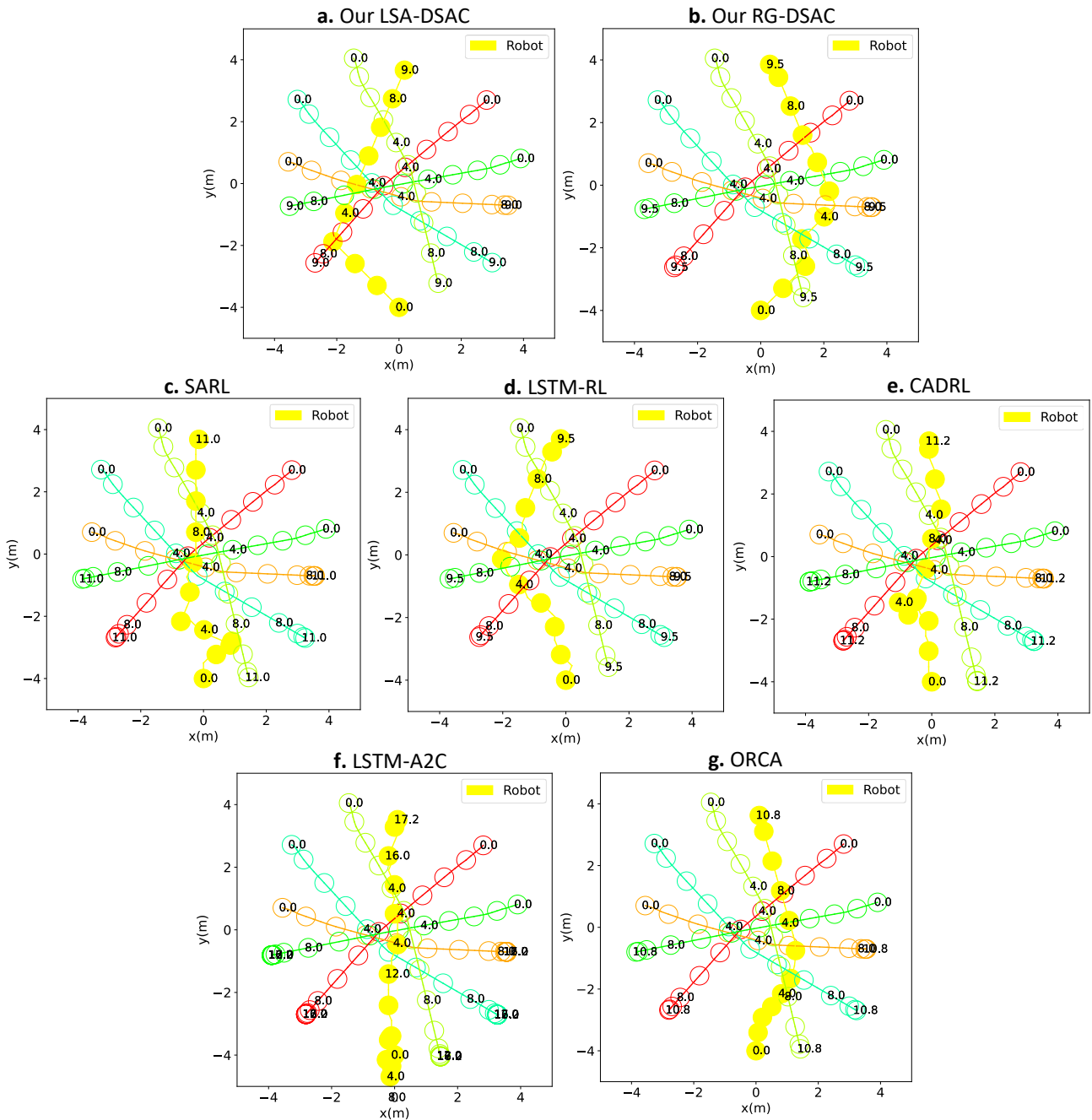

**Fig. 9.** Superiority of LSA-DSAC in trajectory quality. Here this paper presents an example that demonstrates a good performance of our LSA-DSAC in the time cost to reach the goal.

**Qualitative evaluation.** The quality here refers to the trajectory quality of the robot in an episode. In 500 tests, the robot based on our algorithms and state-of-the-art (trainable) learnt some of the six motion planning strategies. These six strategies include *full-bypass*, *partly-bypass*, *follow-pass*, *wait-pass*, *back-pass*, and *cross*. Their features and examples are shown in TABLE IV and Fig. 8. For each algorithm, we sampled 50 trajectories from 500 tests and found that the most of sampled trajectories of the SARL and Our LSA-DSAC followed the high-quality full-bypass and partly-bypass strategies (TABLE V), while the trajectories of the CADRL, LSTMRL, LSTM-A2C, and RG-DSAC followed the medium-quality follow-pass, wait-pass, and back-pass strategies. CADRL and ORCA took low-quality cross strategy that caused more collisions, although the cross strategy led to fast speed sometimes. Fig. 9 presents some examples that indicate the superiority of our LSA-DSAC in trajectory quality when comparing with the state-of-the-art.

**Quantitative evaluation.** The quantity here refers to the statistical motion planning result in 500 tests of each algorithm from the perspectives of the *success rate*, *time to goal*, *collision rate*, *timeout rate* (allowed time 25s), *mean distance of robot and obstacle*, and *mean reward*. The statistics (TABLE VI) shows that our LSA-DSAC outperforms the state-of-the-art in all perspectives, except for the time to goal. However, LSA-DSAC still maintains high performance (2[nd] place) in the time to goal.

TABLE VI. Statistical results of the quantitative evaluation.

| Algorithms | Success rate | Time to goal | Collision rate | Timeout rate | Mean distance | Mean reward |
|---|---|---|---|---|---|---|
| ORCA | 0.43 | 10.86 | 0.564 | 0.006 | 0.08 | --- |
| CADRL | 0.89 | 11.30 | 0.106 | 0.004 | 0.16 | 0.47 |
| LSTMRL | 0.96 | 12.10 | 0.02 | 0.01 | 0.16 | 0.49 |
| SARL | 0.99 | 10.96 | 0.01 | 0.00 | 0.18 | 0.56 |
| LSTM-A2C | 0.88 | 17.04 | 0.05 | 0.07 | 0.12 | 0.36 |
| RG-DSAC | 0.94 | 11.37 | 0.06 | 0.00 | 0.14 | 0.52 |
| LSA-DSAC | **0.996** | 10.94 | **0.004** | **0.00** | 0.15 | **0.57** |

TABLE VII. Time complexity evaluation. The time complexity here is measured by the time cost of all algorithms in training.

| Algorithms | Time cost (hour/10K epi.) |
|---|---|
| ORCA | --- |
| CADRL | 7.4 (train with one obstacle) |
| LSTMRL | 16.08 |
| SARL | 14.72 |
| LSTM-A2C | 0.42 |
| RG-DSAC | 4.38 |
| LSA-DSAC | 4.56 |

TABLE VIII. Transferability evaluation. Here the transferability is measured by the performance of trained models (training in the circle-crossing simulator) in the new environment (the square-crossing simulator). The performance here is measured using the same metrics as that of quantity evaluation.

| Algorithms | Success rate | Time to goal | Collision rate | Timeout rate | Mean distance | Mean reward |
|---|---|---|---|---|---|---|
| ORCA | 0.74 | 9.12 | 0.256 | 0.004 | 0.08 | --- |
| CADRL | 0.88 | 11.19 | 0.01 | 0.11 | 0.17 | 0.48 |
| LSTMRL | 0.91 | 10.54 | 0.03 | 0.06 | 0.12 | 0.49 |
| SARL | 0.92 | 10.96 | 0.02 | 0.06 | 0.17 | 0.51 |
| LSTM-A2C | 0.45 | 15.61 | 0.41 | 0.14 | 0.10 | 0.12 |
| RG-DSAC | 0.40 | 11.09 | 0.59 | 0.01 | 0.11 | 0.10 |
| LSA-DSAC | **0.93** | 10.95 | 0.05 | 0.02 | **0.14** | **0.51** |

TABLE IX. Robustness evaluation. Here the robustness is measured by the value changes of trained models from the circle-crossing simulator to the square-crossing simulator. Robustness evaluation uses the same metrics as that of quantity evaluation.

| Algorithms | Success rate | Time to goal | Collision rate | Timeout rate | Mean distance | Mean rewards |
|---|---|---|---|---|---|---|
| ORCA | 0.310 | 1.560 | 0.308 | 0.002 | 0.000 | --- |
| CADRL | 0.010 | 2.110 | 0.096 | 0.106 | 0.010 | 0.02 |
| LSTMRL | 0.078 | 1.160 | 0.008 | 0.050 | 0.020 | 0.04 |
| SARL | 0.070 | 0.000 | 0.010 | 0.060 | 0.010 | 0.05 |
| LSTM-A2C | 0.430 | 1.430 | 0.360 | 0.070 | 0.020 | 0.24 |
| RG-DSAC | 0.540 | 0.280 | 0.530 | 0.010 | 0.030 | 0.42 |
| LSA-DSAC | 0.066 | 0.010 | 0.046 | 0.020 | 0.010 | 0.06 |

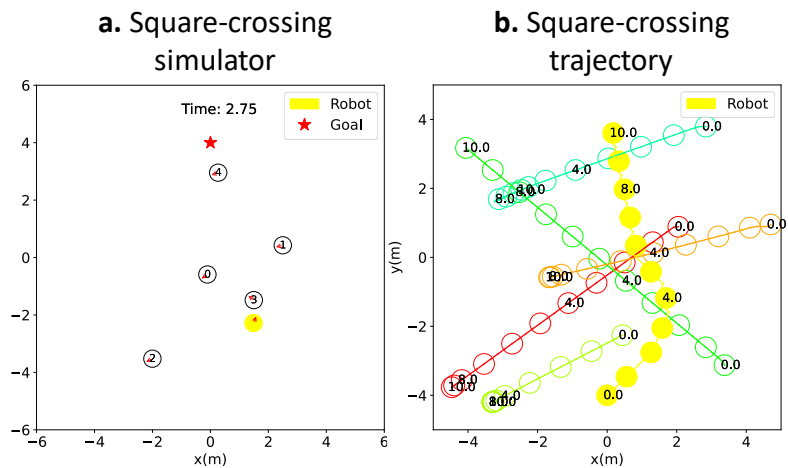

**Fig. 10.** An example of 500 tests in the square-crossing simulator. (a) presents the simulator, while (b) presents the trajectories of the robot and obstacles at the end of an episode. The square-crossing simulator is only used in transferability evaluations.

**Time complexity evaluation.** Here the time complexity is scaled by the *time cost* of each algorithm in training. Online learning algorithm LSTM-A2C learns from online data. Hence, it takes less time 0.42h in training (TABLE VII), while other off-policy algorithms including our RG-DSAC and LSA-DSAC take much time. However, our LSA-DSAC and RG-DSAC still keep high performance (around 4h) when comparing with other off-policy algorithms.

**Transferability evaluation.** The transferability here refers to the performance of a trained model (trained in circle-crossing simulator) in a new environment (square-crossing simulator, Fig. 10). The test results (TABLE VIII) show that our LSA-DSAC keeps the best performance among all trained models in the success rate, mean distance of the robot and obstacle, and mean reward. For the time to goal, collision rate, and timeout rate, our LSA-DSAC still keeps high performance ($3^{rd}$, $4^{th}$, and $3^{rd}$ place respectively).

**Robustness evaluation.** The robustness here denotes the *stability* of a trained model in a new environment. The stability is described as the *value changes* (the changes of the statistical results in the quantitative evaluation). TABLE IX presents the value changes of these models from the circle-crossing simulator to the square-crossing simulator. Although our LSA-DSAC does not perform best among all trained models, it still keeps high performance ($2^{nd}$, $2^{nd}$, $3^{rd}$, $3^{rd}$, $2^{nd}$, and $4^{th}$ place respectively).

*D. Physical implementation*

This paper provides a demonstration of physical implementation. The motivation for physical implementation is to provide a possible way to implement the physical robot and enable the robot to navigate in dense and dynamic scenarios as the motion planning in the simulator. This paper emphasizes the evaluations of motion planning algorithms in simulators, instead of the evaluations in the real world, because simulators can provide as many tests as possible to extensively evaluate the performance of motion planning algorithms. Moreover, the errors introduced by simulators are predictable. In the real world, unexpected errors may result in an unfair environment when evaluating the motion planning performance of algorithms. For example, the false operations from humans and the measurement errors from the sensors may make the results of the real-world evaluations different from that in simulators under the same settings. This paper attempts to create a real-world environment that has the same settings as that of simulators to demonstrate the motion planning performance of algorithms. The unexpected errors from the real world are not considered in the physical implementation. The problems of unexpected errors are expected to be solved by integrating the model-based methods or Bayesian inference into the model-free RL. However, the model-based methods may bring new problems like expensive computation [40]. This paper is not going to extend this topic, but model-based methods may be considered in future works for motion planning in dense and dynamic scenarios.

In the physical implementation, the models of trainable motion planning algorithms are trained with the data from the simulator, while the testing data is collected by the robot sensors from the real world. The mechanism of physical implementation is as follows: A Local area network (LAN) is established in the robot operation system (ROS) [41]. Agents in the experimental area are connected to the LAN. They are equipped with a marker point that can be captured by cameras to create a rigid body model in the motion capture system. The workflow in the physical implementation is shown in Figure 11a. First, cameras capture the agent's location information (observations) via marked points. Second, observations are sent to the host to compute the agent's actions by the RL model and ORCA. Third, the actions are broadcasted to the agents by WiFi. Finally, the actions are executed by agents. Once the robot reaches the goal, the task is finished.

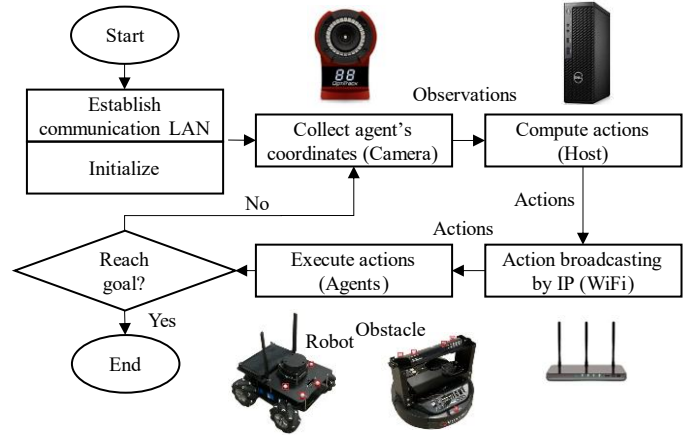

a. Workflow of physical implementation

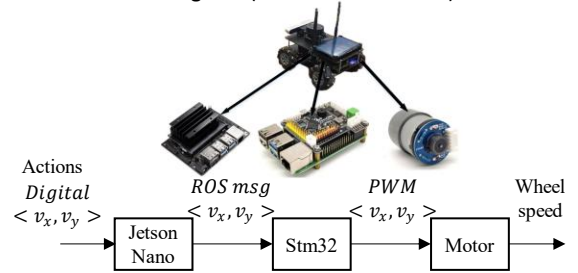

b. Action execution in agents (robot and obstacles)

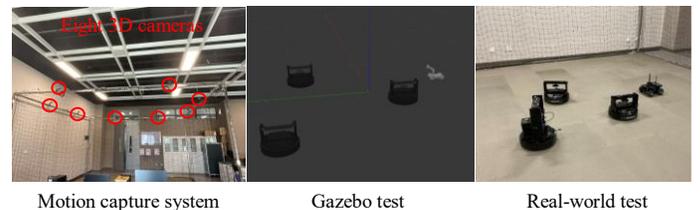

c. Motion capture system, gazebo and real-world tests

**Fig. 11.** Details of physical implementation. (a) presents the detailed steps of physical implementation. (b) presents the hardware of the robot and obstacles for the action execution. (c) presents the motion capture system and an example of the tests in the gazebo and the real world. The motion capture system localizes the robot and obstacles in real-time to compute the positions and velocities of the robot and obstacles.

In action execution (Figure 11b), Jetson Nano converts digital actions to ROS messages. Stm32 then computes the wheel speed using ROS messages and converts it into messages of Pulse-Width Modulation (PWM). They can be recognized and executed by motors. The motion capture system is based on optical tracking technology [42][43][44] to localize agents. It consists of eight 3D cameras. Finally, our physical implementation is tested in both a ROS Gazebo environment and the real world (Figure 11c). Videos of the ROS Gazebo test and the real-world tests are available as follows:

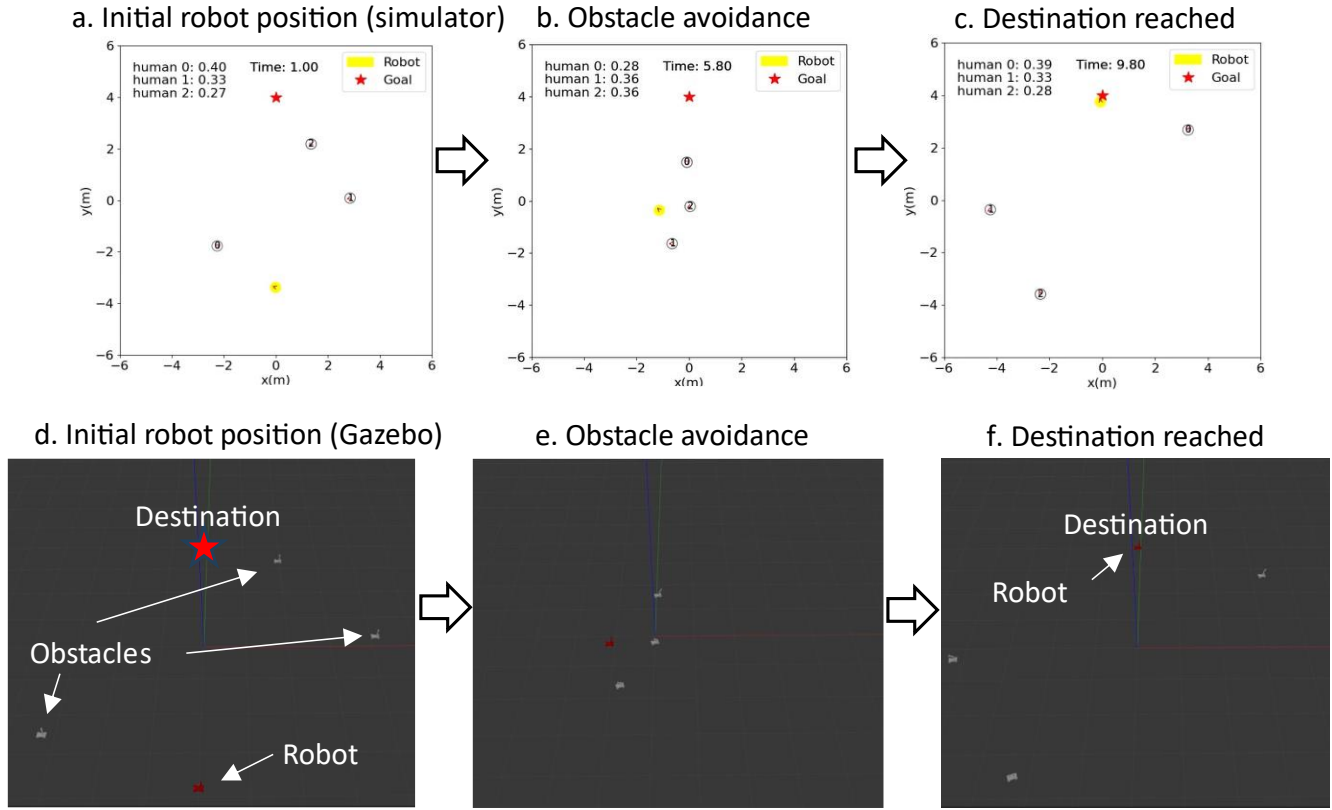

**Fig. 12.** The Gazebo test. The same trained models are evaluated in Gazebo environment and the circle-crossing simulator simultaneously. The robot and obstacles in these two environments are controlled by the LSA-DSAC and ORCA respectively. The model of LSA-DSAC is trained by 1k episodes in the circle-crossing simulator. Finally, the motion planning results in these two environments were almost the same, despite a few differences in the smoothness of the trajectory given the video results.

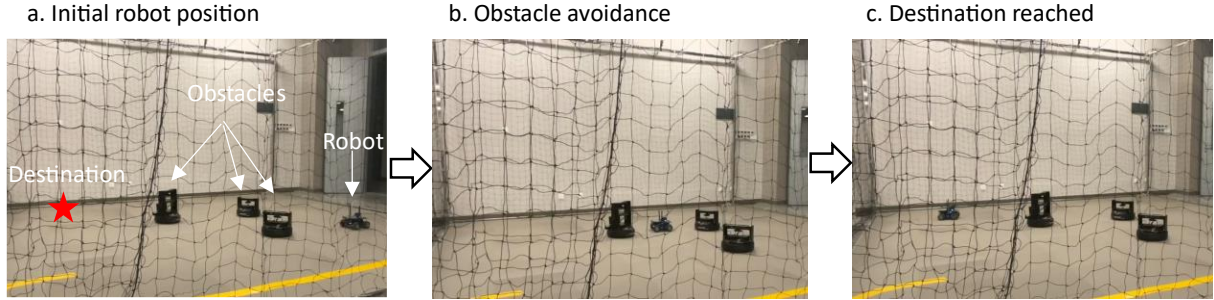

**Fig. 13.** The real-world test in the static environment. As Gazebo test, the real-world test in a static environment uses the model of LSA-DSAC. The model of LSA-DSAC is trained by 1k episodes in the circle-crossing simulator. The obstacles spread along the robot's route to the destination. Finally, the robot reached its destination and avoided all obstacles simultaneously.

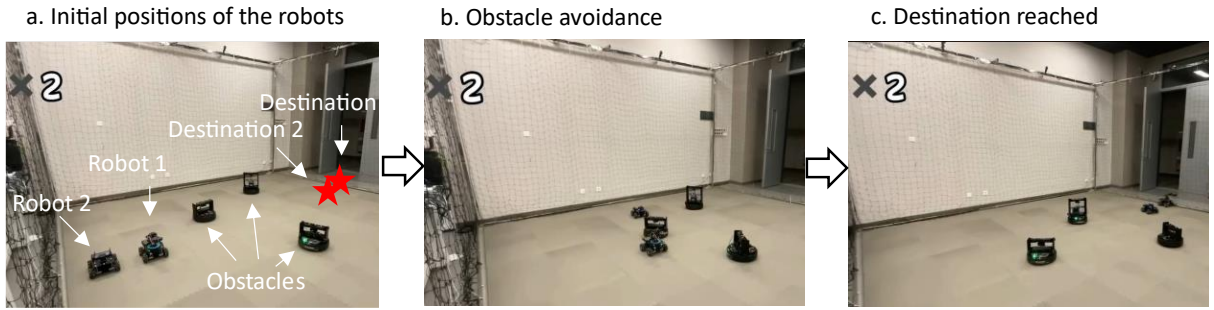

**Fig. 14.** Real-world test in dense and dynamic scenarios. As Gazebo test and real-world test in a static environment, real-world test in dense and dynamic scenarios uses the model of LSA-DSAC. The model of LSA-DSAC is trained by 1k episodes in the circle-crossing simulator. Two robots use the same model of LSA-DSAC. Each robot treats another robot as an obstacle in the motion planning task. The obstacles spread along the robot's routes to their destinations and randomly walk continuously. Finally, the robots reached their destinations and avoided all dynamic obstacles simultaneously. Note that this paper omitted the real-world test in dense and dynamic scenarios with one robot because each robot treats another robot as an obstacle in the motion planning.

1) Gazebo test (Figure 12). The video link is available at https://youtu.be/A-GdHGoWwCk. The purpose of the Gazebo test is to compare the motion planning differences between the Gazebo environment and the simulator. The same trained models are evaluated in Gazebo environment and the circle-crossing simulator. The experiment demonstrates that the motion planning performance in Gazebo environment and circle-crossing simulator is almost the same under the same settings. Their difference is that the trajectories of the robot and the obstacles in Gazebo environment are not as smooth as that of the circle-crossing simulator, given the video demonstration. The sensor errors of Gazebo environment cause the positioning drift, reducing the smoothness of the trajectories. However, the robot can still reach the destination safely and efficiently by learning an efficient motion planning policy and keeping a safe distance to the obstacle.

2) Real-world test in the static environment (Figure 13). The video link is available at the website https://www.youtube.com/watch?v=bH5FbA14AqE. The model tested in Gazebo environment is then tested in the static real-world environment. Given the test result, the robot can reach the destination and avoid all obstacles simultaneously.

3) Real-world test in dense and dynamic scenarios (Figure 14). The video is available at https://youtu.be/UB6aC3XoZ6c. The real-world test in dense and dynamic scenarios uses the same model as the above two tests. Given the test result, the robots can reach their destinations and avoid all dynamic obstacles simultaneously. As Gazebo test, real-world tests in static and dynamic scenarios have the same problem in the positioning drift caused by the sensor errors, reducing the smoothness of the trajectories, given the video demonstration. However, the robot can still reach the destination safely and efficiently by learning an efficient motion planning policy and keeping a safe distance from the obstacle.

## V. CONCLUSION

This paper combines representation learning with reinforcement learning for robotic motion planning in the environment with dynamic and dense obstacles. First, relational graph combines with DSAC to form the RG-DSAC, and satisfactory performance of motion planning is achieved. Second, the expressive power of interpreted features is improved by the attention weight (attention network) to replace the relational graph in the feature interpretation. This improves network convergence. Third, the attention weight (network) is optimized by the skip connection method and LSTM pooling to eliminate overfittings in training. Therefore, the convergence speed and converged result are further improved. Extensive experiments (training and evaluations) of our algorithms and state-of-the-art are conducted. The results demonstrated that our LSA-DSAC outperforms the state-of-the-art in trainings and most evaluations. The details of physical implementation of the robot and dynamic obstacles are also given to provide a possible method to transplant the simulation into the real world. Motion planning experiments were conducted in indoor scenarios (ROS Gazebo environment and real world). This further demonstrates the credibility of our motion planning algorithm and physical implementation method in the real world.

Future research may focus on the design of independent objectives for the attention network to further improve the convergence and interpretability. We will also try tree models and Bayesian model-based method to infer the hidden features of the robot and obstacles. This contributes to better interpretability and reducing unexpected errors once the robot works in the real world, therefore improving the network convergence, and reducing the sim2real gap.

ACKNOWLEDGMENT

The physical implementation is supported in part by the National Natural Science Foundation of China under Grant *62003218*, Guangdong Basic and Applied Basic Research Foundation under Grant *2019A1515110234*, and Shenzhen Science and Technology Program under Grant *RCBS20200714114921371*.

REFERENCES

[1] L. Fuentes-Moraleda, P. Díaz-Pérez, A. Orea-Giner, A. Muñoz- Mazón, and T. Villacé-Molinero, "Interaction between hotel service robots and humans: A hotel-specific Service Robot Acceptance Model (sRAM)," *Tour. Manag. Perspect.*, vol. 36, no. November 2019, p. 100751, 2020.

[2] B. Gerrits and P. Schuur, "Parcel Delivery for Smart Cities: A Synchronization Approach for Combined Truck-Drone-Street Robot Deliveries," *Proc. - Winter Simul. Conf.*, vol. 2021-Decem, no. i, 2021.

[3] S. Srinivas, S. Ramachandiran, and S. Rajendran, "Autonomous robot-driven deliveries : A review of recent developments and future directions," *Transp. Res. Part E*, vol. 165, no. July, p. 102834, 2022.

[4] W. S. Barbosa *et al.*, "Industry 4.0: examples of the use of the robotic arm for digital manufacturing processes," *Int. J. Interact. Des. Manuf.*, vol. 14, no. 4, pp. 1569–1575, 2020.

[5] M. Bartoš, V. Bulej, M. Bohušík, J. Stancek, V. Ivanov, and P. Macek, "An overview of robot applications in automotive industry," *Transp. Res. Procedia*, vol. 55, pp. 837–844, 2021.

[6] P. E. Hart, N. J. Nilsson, and B. Raphael, "A Formal Basis for the Heuristic Determination of Minimum Cost Paths," *IEEE Trans. Syst. Sci. Cybern.*, vol. 4, no. 2, pp. 100–107, 1968.

[7] D. Fox, W. Burgard, and S. Thrun, "The dynamic window approach to collision avoidance," *IEEE Robot. Autom. Mag.*, vol. 4, no. 1, pp. 23–33, 1997.

[8] J. Den Van Berg, M. Lin, and D. Manocha, "Reciprocal velocity obstacles for real-time multi-agent navigation," *Proc. - IEEE Int. Conf. Robot. Autom.*, vol. 2, no. 4, pp. 100–107, 2008.

[9] X. Dai, Y. Mao, T. Huang, N. Qin, D. Huang, and Y. Li, "Automatic obstacle avoidance of quadrotor UAV via CNN-based learning," *Neurocomputing*, vol. 402, pp. 346–358, 2020.

[10] V. Mnih *et al.*, "Playing Atari with Deep Reinforcement Learning," *arXiv*, pp. 1–9, 2013.

[11] V. Mnih *et al.*, "Asynchronous methods for deep reinforcement learning," *33rd Int. Conf. Mach. Learn. ICML 2016*, vol. 4, pp. 2850–2869, 2016.

[12] A. Alahi, K. Goel, V. Ramanathan, A. Robicquet, L. Fei-Fei, and S. Savarese, "Social LSTM: Human Trajectory Prediction in Crowded Spaces," in *2016 IEEE Conference*

[13] C. Chen, S. Hu, P. Nikdel, G. Mori, and M. Savva, "Relational Graph Learning for Crowd Navigation," in *arXiv*, 2019.

[14] W. L. Hamilton, "Graph Representation Learning," *Synth. Lect. Artif. Intell. Mach. Learn.*, vol. 14, no. 3, pp. 1–159, 2020.

[15] C. Chen, Y. Liu, S. Kreiss, and A. Alahi, "Crowd-robot interaction: Crowd-aware robot navigation with attention-based deep reinforcement learning," *Proc. - IEEE Int. Conf. Robot. Autom.*, vol. 2019-May, pp. 6015–6022, 2019.

[16] A. Vemula, K. Muelling, and J. Oh, "Social Attention: Modeling Attention in Human Crowds," *Proc. - IEEE Int. Conf. Robot. Autom.*, pp. 4601–4607, 2018.

[17] H. Van Hasselt, A. Guez, and D. Silver, "Deep reinforcement learning with double Q-Learning," *30th AAAI Conf. Artif. Intell. AAAI 2016*, pp. 2094–2100, 2016.

[18] Z. Wang, T. Schaul, M. Hessel, H. Van Hasselt, M. Lanctot, and N. De Frcitas, "Dueling Network Architectures for Deep Reinforcement Learning," *33rd Int. Conf. Mach. Learn. ICML 2016*, vol. 4, no. 9, pp. 2939–2947, 2016.

[19] D. Silver, G. Lever, N. Heess, T. Degris, D. Wierstra, and M. Riedmiller, "Deterministic policy gradient algorithms," *31st Int. Conf. Mach. Learn. ICML 2014*, vol. 1, pp. 605–619, 2014.

[20] R. Munos, T. Stepleton, A. Harutyunyan, and M. G. Bellemare, "Safe and efficient off-policy reinforcement learning," *Adv. Neural Inf. Process. Syst.*, no. Nips, pp. 1054–1062, 2016.

[21] J. Duan, Y. Guan, S. E. Li, Y. Ren, Q. Sun, and B. Cheng, "Distributional Soft Actor-Critic: Off-Policy Reinforcement Learning for Addressing Value Estimation Errors," *IEEE Trans. Neural Networks Learn. Syst.*, pp. 1–15, 2021.

[22] S. Fujimoto, H. Van Hoof, and D. Meger, "Addressing Function Approximation Error in Actor-Critic Methods," *35th Int. Conf. Mach. Learn. ICML 2018*, vol. 4, pp. 2587–2601, 2018.

[23] T. Haarnoja et al., "Soft Actor-Critic Algorithms and Applications," *arXiv*, 2018.

[24] T. Haarnoja, A. Zhou, P. Abbeel, and S. Levine, "Soft actor-critic: Off-policy maximum entropy deep reinforcement learning with a stochastic actor," *35th Int. Conf. Mach. Learn. ICML 2018*, vol. 5, pp. 2976–2989, 2018.

[25] P. Christodoulou, "Soft Actor-Critic for Discrete Action Settings," *arXiv*, pp. 1–7, 2019.

[26] P. Long, T. Fanl, X. Liao, W. Liu, H. Zhang, and J. Pan, "Towards optimally decentralized multi-robot collision avoidance via deep reinforcement learning," *Proc. - IEEE Int. Conf. Robot. Autom.*, pp. 6252–6259, 2018.

[27] Y. F. Chen, M. Liu, M. Everett, and J. P. How, "Decentralized non-communicating multiagent collision avoidance with deep reinforcement learning," *Proc. - IEEE Int. Conf. Robot. Autom.*, pp. 285–292, 2017.

[28] C. Zhou, B. Huang, H. Hassan, and P. Fränti, "Attention-based advantage actor-critic algorithm with prioritized experience replay for complex 2-D robotic motion planning," *J. Intell. Manuf.*, 2022.

[29] M. Everett, Y. F. Chen, and J. P. How, "Motion Planning among Dynamic, Decision-Making Agents with Deep Reinforcement Learning," *IEEE Int. Conf. Intell. Robot. Syst.*, no. iii, pp. 3052–3059, 2018.

[30] J. Schulman, F. Wolski, P. Dhariwal, A. Radford, and O. Klimov, "Proximal policy optimization algorithms," *arXiv*, pp. 1–12, 2017.

[31] E. Bas, "An Introduction to Markov Chains," *Basics Probab. Stoch. Process.*, pp. 179–198, 2019.

[32] L. Baird, "Residual Algorithms: Reinforcement Learning with Function Approximation," *Mach. Learn. Proc. 1995*, pp. 30–37, 1995.

[33] V. Mnih et al., "Human-level control through deep reinforcement learning," *Nature*, vol. 518, no. 7540, pp. 529–533, 2015.

[34] V. R. Konda and J. N. Tsitsiklis, "Actor-critic algorithms," *Adv. Neural Inf. Process. Syst.*, pp. 1008–1014, 2000.

[35] S. Boyd and L. Vandenberghe, *Convex Optimization*. 2004.

[36] K. He, X. Zhang, S. Ren, and J. Sun, "Deep Residual Learning for Image Recognition," *Proc. IEEE Conf. Comput. Vis. Pattern Recognit.*, pp. 770–778, 2016.

[37] P. Tang, H. Wang, and S. Kwong, "G-MS2F: GoogLeNet based multi-stage feature fusion of deep CNN for scene recognition," *Neurocomputing*, vol. 225, no. November 2016, pp. 188–197, 2017.

[38] G. Huang, Z. Liu, L. van der Maaten, and K. Q. Weinberger, "Densely Connected Convolutional Networks," *Proc. IEEE Conf. Comput. Vis. Pattern Recognit.*, pp. 4700–4708, 2017.

[39] K. Xu, S. Jegelka, W. Hu, and J. Leskovec, "How powerful are graph neural networks?," *7th Int. Conf. Learn. Represent. ICLR 2019*, pp. 1–17, 2019.

[40] C. Zhou, C. Wang, H. Hassan, H. Shah, B. Huang, and P. Fränti, "Bayesian inference for data-efficient, explainable, and safe robotic motion planning: A review," *arXiv:2307.08024*, pp. 1–33, 2023.

[41] M. Quigley et al., "ROS: an open-source Robot Operating System," *ICRA Work. open source Softw.*, vol. 3, no. 3.2, 2009.

[42] J. S. Furtado, H. H. T. Liu, G. Lai, H. Lacheray, and J. Desouza-Coelho, "Comparative analysis of OptiTrack motion capture systems," *Proc. Can. Soc. Mech. Eng. Int. Congr. May 27-30, 2018, Toronto, Canada*, 2018.

[43] T. Ameler et al., "A Comparative Evaluation of SteamVR Tracking and the OptiTrack System for Medical Device Tracking," *Proc. Annu. Int. Conf. IEEE Eng. Med. Biol. Soc. EMBS*, pp. 1465–1470, 2019.

[44] V. García-Vázquez, E. Marinetto, J. A. Santos-Miranda, F. A. Calvo, M. Desco, and J. Pascau, "Feasibility of integrating a multi-camera optical tracking system in intra-operative electron radiation therapy scenarios," *Phys. Med. Biol.*, vol. 58, no. 24, pp. 8769–8782, 2013.